\documentclass{article}

\usepackage{iclr2026_conference,times}

\usepackage{amsmath,amsfonts,bm}

\def\eqref#1{equation~\ref{#1}}

\def\1{\bm{1}}

\DeclareMathAlphabet{\mathsfit}{\encodingdefault}{\sfdefault}{m}{sl}
\SetMathAlphabet{\mathsfit}{bold}{\encodingdefault}{\sfdefault}{bx}{n}

\usepackage{hyperref}
\usepackage{url}

\usepackage[utf8]{inputenc} 
\usepackage[T1]{fontenc}    
\usepackage{booktabs}       
\usepackage{amsfonts}       
\usepackage{nicefrac}       
\usepackage{microtype}      
\usepackage{xcolor}         
\usepackage{comment}
\usepackage{bm}
\usepackage{ulem}
\usepackage{amsmath,amssymb,cases} 
\usepackage{graphicx}

\newcommand{\Msumu}{ M_{\mathrm{sum},u} }

\title{
    Memory Determines Learning Direction: \\
    A Theory of Gradient-Based Optimization in State Space Models
}

\author{
    JingChuan Guan
    \thanks{
        Correspondence to
        : \texttt{kan@isi.imi.i.u-tokyo.ac.jp}},
    \quad
    Tomoyuki Kubota,
    \quad
    Yasuo Kuniyoshi,
    \quad
    Kohei Nakajima \\
    Intelligent Systems and Informatics Laboratory, \\
    Graduate School of Information Science and Technology, \\
    The University of Tokyo
}

\iclrfinalcopy 
\begin{document}

\maketitle

\begin{abstract}
State space models (SSMs) have gained attention by showing potential to outperform Transformers.
However, previous studies have not sufficiently addressed the mechanisms underlying their high performance
owing to a lack of theoretical explanation of SSMs' learning dynamics.
In this study, we provide such an explanation and propose an improved training strategy.
The memory capacity of SSMs can be evaluated by examining how input time series are stored in their current state.
Such an examination reveals a tradeoff between memory accuracy and length,
as well as the theoretical equivalence between the structured state space sequence model (S4) and a simplified S4 with diagonal recurrent weights.
This theoretical foundation allows us to elucidate the learning dynamics, proving the importance of initial parameters.
Our analytical results suggest that successful learning requires the initial memory structure to be the longest possible
even if memory accuracy may deteriorate
or the gradient lose the teacher information.
Experiments on tasks requiring long memory confirmed that extending memory is difficult, emphasizing the importance of initialization.
Furthermore, we found that fixing recurrent weights can be more advantageous than adapting them
because it achieves comparable or even higher performance with faster convergence.
Our results provide a new theoretical foundation for SSMs and potentially offer a novel optimization strategy.
\end{abstract}

\section{Introduction}
In recent years, machine learning models centered around linear recurrent neural networks (RNNs)
have demonstrated strong performance in tasks involving time series modeling~\citep{dao2024transformers}.
These models, derived from control theory, are referred to as state space models (SSMs).
By revealing their learning mechanisms that drive high performance,
many studies have sought to develope faster, more energy-efficient, and more accurate models. 
However, some key techniques introduced in SSMs still lack a solid theoretical foundation. 

A seminal example of this trend are high-order polynomial project operators (HiPPO)~\citep{NEURIPS2020_102f0bb6},
a family of RNNs that achieves a memory architecture theoretically optimized for the longest possible memory, 
based on the Lebesgue measure.
Building on this foundation,
\citet{guefficiently} proposed the S4 (structured state space sequence model)
as a means to implement HiPPO in a computationally efficient way.
These studies showed that the initialization of weights plays a critical role in determining final performance~\citep{NEURIPS2020_102f0bb6, guefficiently};
linear RNNs should be initialized to maximize memory length,
even though doing so may inherently degrade memory accuracy because of a theoretical upper bound~\citep{Jaeger_2001}.
Subsequently, to further reduce computational overhead,
S4D (a simplified S4 with diagonal state matrices)~\citep{gu2022parameterization} was implemented
by simplifying the recurrent weight of the linear RNN into a diagonal matrix.
This technique remains at the core of later developments in SSM research~\citep{dao2024transformers,gu2023mamba}.
Nevertheless, a theoretical explanation justifying these techniques is still lacking.
The reasons why initialization is crucial,
and why such a choice of initialization contributes to performance improvement,
have only been demonstrated empirically.
Moreover, no proof has demonstrated that S4 and S4D are theoretically equivalent.
Prior studies theoretically analyzing the learning dynamics of linear RNNs
have also not solved these questions~\citep{smekal2024towards, proca2024learning, NEURIPS2020_9ac1382f}.
Addressing these issues could lead to better models and more effective training approaches,
which may help mitigate typical machine learning problems,
including overfitting~\citep{hinton2012improving, salman2019overfitting} and long training steps.
To achieve this objective,
we can quantitatively evaluate the information processing characteristics of systems in response to input and compare them across models
using information processing capacity (IPC)~\citep{dambre2012information}.
This indicator statistically measures how a time-evolving information processing system transforms arbitrary inputs
and retains the processed results within the state as memory,
including both linear and nonlinear transformation.
IPC can be used for any information processing system that processes inputs over time,
regardless of system's internal structure.
When calculated with delayed linear inputs, IPC is called memory function (MF)~\citep{Jaeger_2001, guan2025noise, rodan2010minimum, farkavs2016computational, gonon2020memory, ballarin2024memory}.
Because linear RNNs conduct only linear information processing,
we can comprehensively investigate the information processing of linear RNNs using MF,
which serves as a fundamental and effective metric for evaluating information processing properties in linear models.

In this study, we clarify the learning mechanism of SSMs in three stages,
specifically by focusing on S4 and S4D.
First, we reformulate the SSM by retaining only the essential components and demonstrate that MF plays a crucial role in SSMs;
this establishes a foundation for analyzing SSMs through the lens of MF.
Second, by analytically examining the gradients in an SSM layer,
we reveal how the initial structure of the MF influences the learning process;
this highlights the essential role of initialization and shows that,
for successful learning,
the initial MF must prioritize memory length over precision
or crucial label information may be lost during backpropagation through a single SSM layer. 
Combining this theoretical result and the results of MF,
we propose that the linear recurrent weight may not require training
if SSMs are initialized with weights that satisfy the memory requirements.
Finally, we empirically validate our theoretical results by evaluating performance across all tasks
in the long range arena (LRA) benchmark~\citep{tay2020long}
and demonstrate that our proposed method is effective.
The results show faster convergence, mitigation of overfitting,
and the potential for further improvements in accuracy.

\section{Background}
\subsection{Structured state space sequence model}
S4 is a deep learning architecture designed to efficiently capture long-range dependencies in sequential data.
It builts on continuous-time linear SSMs,
particularly those formulated in the HiPPO framework,
and translates them into trainable and hardware-efficient architecture suitable for modern sequence modeling tasks.
S4 begins with the general form of a continuous-time SSM:
\begin{align}
\begin{array}{ccc}
    \dot{\bm{x}}(t) = \tilde{\bm{A}} \bm{x}(t) + \tilde{\bm{B}} \bm{u}(t)
    & &
    \bm{y}(t) = \tilde{\bm{C}} \bm{x}(t) + \tilde{\bm{D}} \bm{u}(t), 
    \label{eq:continuous SSM}
\end{array}
\end{align}
where $ \bm{x}(t) \in \mathbb{R}^N $ is the state, $ \bm{u}(t) \in \mathbb{R} $ is the input, and $ \bm{y}(t) \in \mathbb{R} $ is the output.
The matrices $ \tilde{\bm{A}} \in \mathbb{R}^{N \times N} $, $ \tilde{\bm{B}} \in \mathbb{R}^{N \times 1} $, $ \tilde{\bm{C}} \in \mathbb{R}^{1 \times N} $, and $ \tilde{\bm{D}} \in \mathbb{R} $ define the system dynamics and output behavior.
In a prior study, HiPPO matrices were adopted as $\tilde{\bm{A}}$ and $\tilde{\bm{B}}$, theoretically realizing the longest memory structure~\citep{gu2020hippo}.
In S4D, ``S4Dinv'' and ``S4Dlin'' were introduced by approximating the eigenvalues of the HiPPO matrices~\citep{gu2022parameterization} (see Appendix~\ref{appendix:model}).
For practice, the continuous model (Eq.~\ref{eq:continuous SSM}) is discretized as follows:
\begin{align}
\begin{array}{ccc}
    \bm{x}_{k+1} = \bm{A} \bm{x}_k + \bm{B} \bm{u}_k
    & &
    \bm{y}_k = \tilde{\bm{C}} \bm{x}_k + \tilde{\bm{D}} \bm{u}_k,
    \label{eq:discrete SSM}
\end{array}
\end{align}
where $\bm{x}_k$, $\bm{u}_k$, and $ \bm{y}_k$ denote the state, input, and output at discrete time $ k $, respectively.
The detailed discretization procedure of the variables and parameters $\bm{A}$ and $\bm{B}$ is described in Appendix~\ref{appendix:model}.
The S4 architecture extends the basic SSM by enabling both parallelization and deep stacking across multiple layers.
In the parallel setting,
a sequence of inputs $ \{ \bm{u}_k \}_{k=1}^T $ is processed by applying the discretized state-space recurrence to each element independently using fast convolution methods,
yielding an output sequence $ \{ \bm{y}_k \}_{k=1}^T $.
To construct a deep S4 model, multiple such layers are stacked, 
with each layer consisting of a structured SSM followed by a pointwise nonlinearity and a linear transformation.
Let $ \bm{u}_{k}^{(\ell)} $ denote the output at time step $k$ in layer $\ell$.
Then, the $\ell$th layer of the S4 network is defined by
\begin{align}
\begin{array}{ccc}
    \bm{x}_{k+1}^{(\ell)} = \bm{A}^{(\ell)} \bm{x}_k^{(\ell)} + \bm{B}^{(\ell)} \bm{u}_{k}^{(\ell-1)}
    & &
    \bm{u}_{k}^{(\ell)} = \bm{W}^{(\ell)} f ( \bm{y}_k^{(\ell)} ) + \bm{b}^{(\ell)} ,
\end{array}
\end{align}
where
$ (\cdot)^{(\ell)} $ denotes the value in $ \ell $th layer,
$ f(\cdot) $ is a nonlinear activation function [e.g., GELU or GLU~\citep{hendrycks2016gaussian, dauphin2017language}],
and $ \bm{W}^{(\ell)} $ and $ \bm{b}^{(\ell)} $ are the weight and bias parameters, respectively, of a linear transformation applied after the state-space computation.

\subsection{Memory capacity}
We evaluate the information processing in the discretized SSM using MF~\citep{Jaeger_2001, dambre2012information},
which represents how well the delayed input series $u_{t-\tau}$ can be reconstructed from the current network state.
Following previous studies, we adopt the uniformly random input $u_t\in[-1, 1]$ to measure MF. 
The emulation is conducted by a linear approximation:
$\hat{u}_{t-\tau} = \bm{w}_{\text{out}}^{\top} \bm{x}_t$,
where $\tau$ is the delay from the current time.
The readout weight vector $\bm{w}_{\text{out}}$ is determined by minimizing a loss function of the mean squared error (MSE):
$L_{\text{MSE}}=\frac{1}{T}\sum_{t=1}^T (\hat{u}_{t-\tau}-u_{t-\tau})^2$,
where $T$ is the sampled time length.
Accordingly, the MF is defined as follows:
\begin{eqnarray}
M[u_{t-\tau}] = 1 -
\frac{
\text{min}_{\bm{w}_{\text{out}}} \langle (\hat{u}_{t-\tau}-u_{t-\tau})^2 \rangle
}{\langle u_{t-\tau}^2\rangle} ~(\leq 1),
\label{eq:original MF}
\end{eqnarray}
where
$ \langle \cdot \rangle$ denotes the time average.
The upper bound $1$ is satisfied when the system has fully memorized the target,
which is the delayed input series.
The sum of the MF with respect to all $\tau$ represents the memory capacity (MC) of the SSM:
\begin{align}
\Msumu =\sum_{\tau=0}^\infty M[u_{t-\tau}] ~(\leq N).
\label{eq:MC,sumMF}
\end{align}
The upper bound of $\Msumu$ is determined by the number of linearly dependent time series in the state,
which is called the rank and is ideally $N$.
This implies that there is a tradeoff between the accuracy and length of memory.
By calculating Eq. (\ref{eq:original MF}),
\citet{dambre2012information} introduced another expression for the MF:
$
M[u_{t-\tau}] =
\bm{U}_{\tau}^{\top}\bm{X}(\bm{X}^{\top}\bm{X})^{-1}\bm{X}^{\top}\bm{U}_{\tau}/(\bm{U}_{\tau}^{\top} \bm{U}_{\tau}),
$
where $ \bm{X}\in\mathbb{R}^{T\times N} $ is a matrix whose columns represents the state time series,
and $\bm{U}_{\tau}\in\mathbb{R}^{T}$ is the delayed input series.
The MF and MC of linear RNNs were analyzed in a previous study~\citep{guan2025noise}:
\begin{align}
\begin{array}{ccc}
    M[u_{t-\tau}]
    =
    {\bm{V}_{\tau}}^{\top}
    ( \bm{V} \bm{V}^{\top} )^{-1}
    {\bm{V}_{\tau}}
    & &
    \Msumu
    =
    \textrm{tr}\left[
    {\bm{V}}^{\top}
    ( \bm{V} \bm{V}^{\top} )^{-1}
    {\bm{V}}\right],
    \label{eq:analytical MF}
\end{array}
\end{align}
where
$\bm{V}=
\begin{pmatrix}
    \bm{V}_{T-1} &  \cdots & \bm{V}_{1} & \bm{V}_{0}
\end{pmatrix}
$ is a Vandermonde matrix,
$\bm{V}_{\tau}=
\begin{pmatrix}
    {\lambda_1}^{\tau} & {\lambda_2}^{\tau} & \cdots & {\lambda_N}^{\tau}
\end{pmatrix}^\top
$,
and 
$ \{ \lambda_i \}_{i=1}^N $ are the eigenvalues of matrix $\bm{A}$ in Eq. \ref{eq:discrete SSM}.
The maximum absolute eigenvalue is called the spectral radius,
which is a crucial parameter for the dynamics of linear RNNs.
If this value is greater than $1$,
the states will diverge as the time step progresses.

\subsection{Gradient based learning}
One of the most widely used parameter update strategies in machine learning models is based on the principle of gradient descent.
Variants such as stochastic gradient descent (SGD)~\citep{robbins1951stochastic, bottou1991stochastic} and Adam~\citep{kingma2014adam} exist.
In all of these methods, errors propagate from the final layer backward,
and the gradients computed at each layer determine the direction of parameter updates.
Here, the error at the final layer is defined as $L_o(\hat{\bm{y}}_o, \bm{y}_o)$,
where $ \hat{\bm{y}}_o \in \mathbb{R}^{H_o}$ is the model output,
$ \bm{y}_o \in \mathbb{R}^{H_o}$ is the target label,
and $H_o$ is the dimensionality of the desired output.
In gradient-based learning,
to evaluate the error of the previous layer and determine the parameter update direction,
the gradient $\nabla_{\bm{\theta}_o} L_o(\hat{\bm{y}}_o, \bm{y}_o)$ of the final layer is calculated,
where $\bm{\theta}_o$ denotes the parameters of the output layer. 
Subsequently, the gradient for the layer immediately preceding the output layer is computed,
a process that is applied recursively across layers and forms the basis for parameter updates.
Accordingly, the optimal output of $l$th layer is determined by the gradient passed from the $(l+1)$th layer,
suggesting that we can define the loss function $L_l(\hat{\bm{y}}_l, \bm{y}_l)$ for the $l$th layer in a manner similar to the output layer,
where $\hat{\bm{y}}_l$ and $\bm{y}_l$ represent the layer output and the ideal outputs of that layer.

In this study, we assume that the loss function $L_l$ of the SSM layer can be represented by the MSE.
Accordingly, the gradient of the SSM is described by
\begin{eqnarray}
\nabla_{\bm{\theta}_l} L_l(\hat{\bm{y}}_l, \bm{y}_l) = \nabla_{\bm{\theta}_l}\frac{1}{H_l} ||\hat{\bm{y}}_l - \bm{y}_l||^2,
\end{eqnarray}
where $\bm{\theta}_l$ denotes the parameters of $l$th layer
and $H_l$ is the dimensionality of the layer output.

\section{Results}
\subsection{MF of SSM}\label{sec:MF of SSM}
In this section, we show why MF is an effective indicator for SSMs in two steps.
First, we analytically simplify the SSM based on the theoretical results of MF obtained in previous studies.
Second, we numerically demonstrate the MF of SSM architectures proposed by prior research
and explain their memory profiles in terms of MF.

\textbf{S4D is theoretically equivalent to S4}\quad
According to a previous study~\citep{dambre2012information},
all information processing of linear SSMs can be comprehensively measured by MF.
The state $\bm{x}_t$ of linear RNNs can be described by
$\bm{x}_t = \sum_{\tau=0}^\infty \bm{\alpha}_{\tau}u_{t-\tau}$,
where $\{\bm{\alpha}_{\tau}\}_{\tau=0}^{\infty}$ is a certain series of fixed constant vectors.
This indicates that the output $y_t$ from the readout is represented by $y_t=\sum_{\tau=0}^\infty \alpha_{\tau}u_{t-\tau}$
using a constatnt sequence $\{\alpha_{\tau}\}_{\tau=0}^{\infty}$,
suggesting that linear RNNs perform information processing solely by computing a linear combination of past inputs.
Since the MF $M[u_{t-\tau}]$ is statistically equivalent to the norm of $\bm{\alpha}_{\tau}$,
it can fully capture the information processing characteristics of linear SSMs.
Combining this result with the analytical solution obtained from \citet{guan2025noise} (Eq.~\ref{eq:analytical MF})
reveals that processed inputs held in the SSM are determined only by the eigenvalues of matrix $\bm{A}$ (Eq.~\ref{eq:discrete SSM}).
This proves that both the input weight $\bm{B}$ and the eigenvectors of $\bm{A}$ can be freely specified,
demonstrating that S4 is theoretically equivalent to S4D because they share the same eigenvalues.
Accordingly, we redefine the recurrent equation of the discrete SSM:
\begin{align}
\bm{x}_{k+1} &= \bm{\Lambda} \bm{x}_k + \bm{B} \bm{u}_k, \label{new discrete SSM}
\end{align}
where
$\bm{\Lambda}
=\text{diag}\begin{pmatrix}
\lambda_1 & \lambda_2 & \cdots& \lambda_N
\end{pmatrix}
$.
Unless otherwise specified, we use input weight $ \bm{B}=B\mathbf{1}$,
where
$ \mathbf{1} \in \mathbb{R}^{N \times 1}$ is the vector whose elements are all $1$
and $B$ is an arbitrary value.

\begin{figure}[t]
\centering
\includegraphics[width=1.0\linewidth]{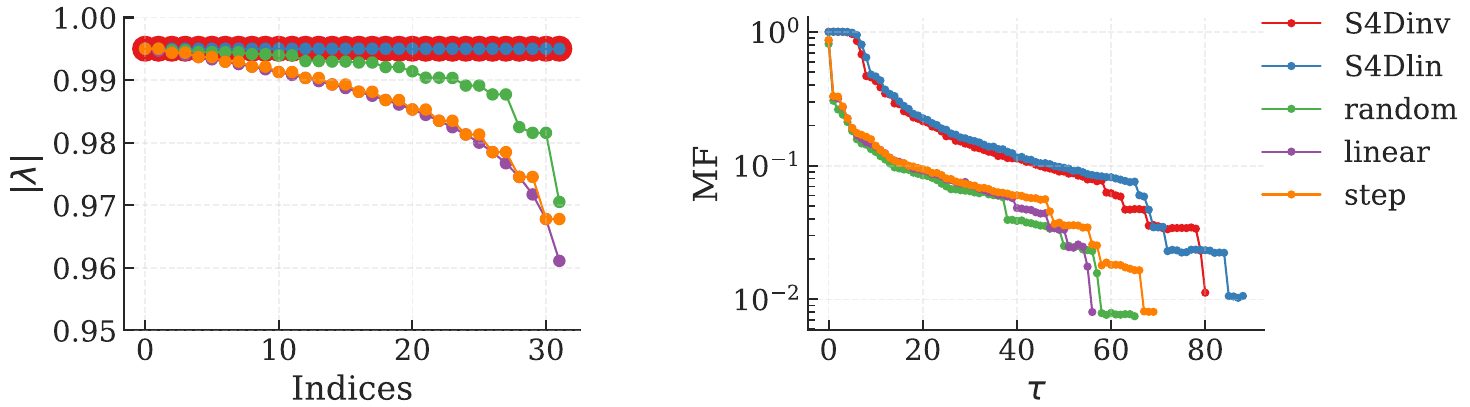}
\caption{\label{fig:MF_S4}
The MF of five eigenvalue realizations.
The left (right) panel shows the absolute eigenvalues (MF).
The horizontal axis is the index of eigenvalues (delay $\tau$ of input).
The system size $N$ is $32$, and the results were obtained by calculating Eq.~\ref{eq:original MF},
where $T=1024$ was typically used in previous studies' models.
}
\end{figure}

\textbf{MF of HiPPO eigenvalues}\quad 
To demonstrate the effectiveness of MF,
we calculated it for five typical eigenvalue realizations and analyzed the corresponding memory structures.
Two groups of eigenvalues for the SSMs are employed.
The first group consisted of those proposed in previous studies as realizing the longest memory structure---namely,
``S4Dinv'' and ``S4Dlin.''
The second group consisted of other representative eigenvalues:
``random'', ``lin'', and ``step.''
``random'' refers to eigenvalues of matrix $\bm{A}$ randomly generated from a uniform distribution.
``lin'' indicates real eigenvalues that are completely linearly distributed.
``step'' indicates that all eigenvalues are complex conjugate pairs whose magnitudes are arranged in a linear distribution.
We evaluated the MFs based on the actual setting used for training and evaluation of the model.
In principle, one SSM layer consists of multiple distinct SSMs.
To evaluate the information processing of the entire SSM,
we defined the supremum MF of the $l$th layer:
\begin{align}
M[u_{t-\tau}]^{(\ell)} = \sup M[u_{t-\tau}]^{(\ell)}_i = \max_{i} M[u_{t-\tau}]^{(\ell)}_i, \label{eq:MF sup}
\end{align}
where $i$ is the index indentifying a certain SSM.
Figure~\ref{fig:MF_S4} shows the eigenvalues and the MF of one layer (Eq.~\ref{eq:MF sup}),
and it illustrates that
``S4Dinv'' and ``S4Dlin'' exhibited longer and stronger memory than other eigenvalue realizations,
suggesting they possess richer information processing.
Clearly, the sums of MFs differ.
This can be explained by the fact that the MF sum numerically does not reach the upper bound, number of nodes $N$,
and the system ranks are different between systems defined by the eigenvalues.
Therefore, the longer MFs are attributed to the characteristics of eigenvalues; ``S4Dinv'' and ``S4Dlin''
inherently generate higher system ranks.
From the perspective of the MF,
this eigenvalue group can be considered superior to the other group.
We provide the motivation for choosing these configurations and the detailed procedure of analysis
in Appendices~\ref{appendix:model} and \ref{appendix:MF of SSM}.

\subsection{Importance of initial parameters for successful learning}
\label{sec:Importance of initial parameters}
Next, we investigate the learning process by focusing on the parameter update method.
To simplify the problem, we introduce two assumptions:
(i) the input $u_t$ is sampled randomly from a uniform distribution, consistent with the setting used for the MF;
(ii) we consider the case where matrix $\bm{A}$ is fixed at its initial value.
Assuming that the SSM has an ideal output series $\bm{y} = \{ y_i\}_{i=1}^T$,
the loss function of single SSM layer is described by
$L_{\text{MSE}}
= \frac{1}{T}\sum_{i=1}^{T} (f(\bm{\theta}^\top \bm{x}_i) - y_i)^2$,
where $T$ is the output dimension of SSM, $\bm{\theta}=\bm{W}^{(\ell)}$ is the parameter of the linear readout layer,
$f(\cdot )$ is the piecewise activation function applied after the linear readout,
and $\bm{x}_i$ is the states used to produce each $y_i$.
The gradient $G(\bm{y})$ can be computed as 
\begin{eqnarray}
G(\bm{y})
=
\nabla_\theta
L_{\text{MSE}}
=
\frac{2}{T}\bm{X}^{\top}
\left[( f(\bm{X}\bm{\theta} ) - \bm{y} )
\odot f'(\bm{X}\bm{\theta} )\right].
\label{nonlinear: Du def}
\end{eqnarray}
To analyze what affects the parameter update process intuitively,
we focus on the SSM before the nonlinear readout is applied.
In the case where $f$ is linear, the gradient is described as follows:
\begin{eqnarray}
G(\bm{y}) =
\frac{2}{T} \bm{X}^{\top}(\bm{X}\bm{\theta} - \bm{y}).
\label{linear: Du def}
\end{eqnarray}
Following the approach of the MF,
we consider a task to predict a delayed input series $\bm{y}=\bm{U}_{\tau}\in\mathbb{R}^{T}$.
The gradient in this task is computed as
$
G(\bm{U}_{\tau})=
2\langle u^2\rangle B\bm{V} (
    B\bm{V}^{\top}\bm{\theta}
    - \bm{V}^{\top}(\bm{V}\bm{V}^{\top})^{-1}\bm{V}_{\tau}
)
$.
Using the fact that linear SSM performs only linear information processing,
the linear combination of past input series $
\bm{y}=\sum_{k=0}^{T-1}\alpha_k\bm{U}_k
$ fully describes the possible outputs of the SSM,
where $\{\alpha_{\tau}\}_{\tau=0}^{T-1}$ is a series characterizing $\bm{y}$.
Accordingly, the gradient $G(\bm{y})$ calculated with this output $\bm{y}$
describes all possible gradients of the SSM layer.
The gradient is computed as
\begin{eqnarray}
G(\bm{y}) =
2\langle u^2\rangle B\bm{V}(
    B\bm{V}^{\top}\bm{\theta}
    - \bm{V}^{\top}(\bm{V}\bm{V}^{\top})^{-1}\bm{V}\bm{\alpha}
).\label{linear: Du with MF}
\end{eqnarray}
The derivation of these formulas is shown in Appendix~\ref{appendix:Parameter Update Algorithm of SSMs}.
According to Eq. \ref{linear: Du with MF}, two important insights are revealed.
First, the state series $\bm{X}$ (Eq. \ref{linear: Du def}) is converted to the matrix $B\bm{V}$,
which suggests that $\bm{V}$ essentially represents the state series during parameter updates.
Second, the matrix $\bm{V}^{\top}(\bm{V}\bm{V}^{\top})^{-1}\bm{V}$,
which is the normalized version of $\bm{V}$, acts as a weighting function for $\bm{\alpha}$.
The $\tau$th element of the vector $\bm{\alpha}$ is multiplied with the $\tau$th column of $\bm{V}$.
The $\tau$th column of $\bm{V}$, denoted as the vector $\bm{V}_{\tau}$,
represents the decay of input information from $\tau$ steps ago.
From this, it is theoretically shown that the decay of past input information in SSMs determines
how much information from $\tau$ steps ago affects the parameter update.
Because the decay of input information stored in the state time series can be evaluated by the MF,
and because $\bm{V}$ represents $\bm{X}$ in the gradient computation,
the decay represented by $\bm{V}$ can be evaluated through the MF.
In fact, the diagonal components of the matrix $\bm{V}^{\top}(\bm{V}\bm{V}^{\top})^{-1}\bm{V}$ are the MF.
Therefore,
if the teacher time series $\bm{y}$ contains past input information whose memory is evaluated as $0$ through MF,
such information is not reflected in the learning process.


\textbf{Memory requirement that must be realized at initialization}\quad
Consider an extreme case where $\bm{y}$ is solely composed of inputs from the distant past such that the MF indicates approximately $0$.
In this case,
because
\begin{eqnarray}
\bm{V}^{\top}(\bm{V}\bm{V}^{\top})^{-1}\bm{V}\bm{\alpha} \approx \bm{0}, 
\end{eqnarray}
the parameter update is independent of the desired output,
the learning process ideally does not proceed.
As there is no explicit mechanism to update the MF through the learning, 
the initialization of eigenvalues is crucial.
To ensure an update reflecting the teacher information,
we must initialize the weight with an MF that meets the following condition:
The values of MF remain sufficiently larger than $0$
in the distant past such that all $\alpha_k$ are not $0$.
The result reveals a crucial requirement for SSMs to solve universal tasks.
To learn tasks that rely on information from arbitrarily distant past inputs,
the SSM must be initialized with an infinitely long memory.
According to the tradeoff between the accuracy and length of memory (Eq.~\ref{eq:MC,sumMF}),
we must satisfy this condition by abandoning the accuracy of memory to some extent.
Our result suggests that this condition may be given higher priority over the memory accuracy
because gradient updates are still executed using teacher signals,
even when the memory is very weak.
However, if accuracy is prioritized,
teacher signals from the distant past are likely to be completely discarded,
and effective updates may not progress at all. 
The meaning of the product $\bm{V}^{\top}(\bm{V}\bm{V}^{\top})^{-1}\bm{V}\bm{\alpha}$
is further analyzed in Appendix~\ref{appendix:importance of MF}.

\textbf{SSM with nonlinear output layer}\quad
The analysis restricted to the linear readout can fully account for the nonlinear readout case
because the input of readout layer is entirely dependent on the output of SSM.
However, to illustrate how nonlinearity affects the output of the full SSM layer, 
we provide the formula for nonlinear readout here.
In the case where $f$ is nonlinear, the gradient (Eq.~\ref{nonlinear: Du def}) is described by
\begin{eqnarray}
G(\bm{y})
&=&
\frac{2}{B}\bm{X}^{\top}
\left[
    f(\bm{X}\bm{\theta} )\odot f'(\bm{X}\bm{\theta} )
    -
    \bm{\hat{U}}^{\top}
    \bm{V}^{\top}(\bm{V}\bm{V}^{\top})^{-1}\bm{V}
    \bar{\bm{\alpha}}
\right],
\end{eqnarray}
where $
\bm{\hat{U}} =
\begin{pmatrix}
    \bm{U}_{T-1} & \cdots & \bm{U}_{0}
\end{pmatrix}$.
The vector $\bar{\bm{\alpha}}$ is the series
characterizing the label $\bar{\bm{y}} = \bm{y} \odot f'(\bm{X}\bm{\theta})$,
modified by the derivative of the activation function,
assuming the input $u_t$ can be considered the source of $\bm{y}$.
We can confirm that 
the update direction is determined by the vector 
$\bm{V}^{\top}(\bm{V}\bm{V}^{\top})^{-1}\bm{V} \bar{\bm{\alpha}}$.
Therefore, we can elucidate the update direction by the product of MF and the modified label series similarly.
The same result applies in the nonlinear case. 

\textbf{Training eigenvalues}\label{sec:Training eigenvalues}\quad
The risk of learning is implied through our analysis.
Here we remove assumption (ii)
and allow the SSM layer to learn the eigenvalues of the internal weight matrix.
In this case, the MF changes during the training process,
and the memory within certain time regions may be lost
because the sum of the MF is theoretically constant (Eq.~\ref{eq:MC,sumMF}).
To increase the MF within a specific time region,
it is theoretically necessary to reduce the MF in other regions. 
Even numerically, 
we cannot ensure that the lost memory will not be required again during training.
For example, when we use the structured eigenvalues [e.g., ``S4Dinv'' and ``S4Dlin''~\citep{gu2022parameterization}],
their theoretically longest memory structure can be destroyed.
On the other hand, 
the advantages of learning eigenvalues can also be explained using MF
in terms of both shortening and extending memory.
Shortening the MF is beneficial in tasks where only short memory is required.
It is preferable to concentrate on essential memory rather than sustain low-precision memory over long delays, 
because the sum of the MF is theoretically fixed. 
The benefits of extending the MF have already been discussed above.
As already clarified,
gradient-based learning does not possess any explicit mechanism to deliberately extend the MF in regions where no memory is present.
Nevertheless, as eigenvalues are adapted to alter the organization of existing memory,
an unintended extension of the MF may also occur. 
For example, when the MF at $\tau=80$ is enhanced during learning with eigenvalues of ``S4Dinv'' in Figure~\ref{fig:MF_S4},
the MF at $\tau=100$, which is initially $0$, may also emerge, though this is not guaranteed by the existing algorithm.
To determine which learning strategy is more effective, numerical experiments are conducted in the next section.

\subsection{SSM with fixed eigenvalues}\label{sec:SSM with fixed eigenvalues}
To address the question raised in the previous section,
we propose the use of an SSM with fixed eigenvalues.
We refer to this setup as the reservoir computing (RC) setting,
following the RC approach in which internal layer weights remain fixed and only output weights are learned
~\citep{jaeger2001echo, maass2002real, lukovsevivcius2009reservoir, nakajima2021reservoir}.
This approach not only preserves the memory structure but also addresses common issues encountered in machine learning,
such as overfitting owing to a large number of trainable parameters,
potentially impairing generalization performance.
To demonstrate the effectiveness of this learning method
and to investigate how initializing with the longest memory structure affects learning,
we evaluate all five eigenvalue initializations for the SSM.
As benchmark tasks, we employed the LRA benchmark,
which is widely used in prior work~\citep{guefficiently, gu2022parameterization} and requires long-term memory to solve.
For all tasks and under each eigenvalue realization,
we computed task accuracies in two settings:
(1) a setting in which eigenvalues are allowed to be learned and (2) the RC setting.
To examine the robustness of each model, we trained them with six different random seeds.
Table~\ref{table:acc} presents the maximum and minimum accuracies obtained under each condition.
Except for the random seed,
all model hyperparameters were kept identical to those used in previous work~\citep{guefficiently}.
Detailed values are provided in Appendix~\ref{appendix:Experimental Details}.

\begin{table}[tbh]
  \caption{
    \label{table:acc}
    Test accuracy of S4D on LRA tasks with different eigenvalue initializations in the SSM.
    Each task was executed six times with different random seeds to investigate whether the models were robust. 
    The minimum and maximum accuracies across runs are reported. 
    For reference, we include the S4 results reported in the official repository~\citep{guefficiently}.
    Three settings are presented: ``S4'', ``learnable eigenvalues'', and ``RC.'' 
    Accuracies are underlined when a given eigenvalue realization achieves the highest accuracy across each setting.
    In addition, when ``RC'' surpasses ``learnable eigenvalues,''
    the corresponding accuracy is bolded. 
    }
  \centering
  \resizebox{\textwidth}{!}{%
  \begin{tabular}{@{}l ll ll ll ll ll ll @{}}
    \toprule
    \midrule
    & \multicolumn{2}{c}{Listops} & \multicolumn{2}{c}{Text} & \multicolumn{2}{c}{Retrieval} & \multicolumn{2}{c}{Image} & \multicolumn{2}{c}{Pathfinder} & \multicolumn{2}{c}{PathX}\\
    Eigenvalues & max acc & min acc & max acc & min acc & max acc & min acc & max acc & min acc & max acc & min acc & max acc & min acc \\
    \midrule
    \midrule
    S4          & 0.595         && 0.865        && 0.910    && 0.885         && 0.940            && 0.960
    \\
    \midrule
    S4Dinv      & 0.594        & \uline{0.589}& 0.853     & 0.839     & 0.863     & 0.760   & 0.864         & 0.859         & \uline{0.910}     & \uline{0.877} & 0.922             & 0.903
    \\
    S4Dlin      & 0.598        & 0.587        & 0.848     & 0.830     & 0.843     & 0.750   & \uline{0.868} & \uline{0.860} & 0.896             &0.865          & \uline{0.926}     & \uline{0.912}
    \\
    random      & 0.552         &0.476        & 0.857     & 0.835     & \uline{0.903}     & \uline{0.899}   & 0.820       & 0.815         & 0.870             & 0.846            & 0.877          & 0.841
    \\
    linear      & 0.543         &0.488        & 0.839     & 0.811     & 0.899     & 0.894   & 0.782       & 0.768         & 0.797             & 0.754            & 0.820          & 0.503
    \\
    step        & \uline{0.605}& 0.575     &\uline{0.862} &\uline{0.852}    & 0.883     & 0.841     & 0.789         & 0.779         & 0.858             &0.824         & 0.840             & 0.824
    \\
    \midrule
    RC S4Dinv      & 0.585         & 0.570         & \textbf{\uline{0.857}} & \textbf{\uline{0.846}}  & \textbf{\uline{0.906}} & \textbf{\uline{0.901}}  & \textbf{0.873}       &\textbf{0.865}         & \uline{0.899} &\textbf{\uline{0.889}} & \textbf{0.932}            & \textbf{\uline{0.923}}
    \\
    RC S4Dlin      & \uline{0.593} & \uline{0.577} & \textbf{0.850}         & \textbf{0.843}            & \textbf{0.903}        & \textbf{0.898}       & \textbf{\uline{0.877}} &\textbf{\uline{0.871}} & 0.892        &\textbf{0.872}          &   \textbf{\uline{0.933}}  & \textbf{0.922}
    \\
    RC random      & 0.535      &\textbf{0.477}        & 0.845     & 0.813     & 0.901     &0.896       &0.733      &0.698         & 0.824          & 0.777             & 0.820             & 0.773
    \\
    RC linear      & 0.530      &0.423        & 0.815     & 0.801     & 0.894     &0.889    &0.686      &0.634         & 0.769          & 0.506             & 0.669             & \textbf{0.504}
    \\
    RC step        & 0.556        & 0.500        & 0.845     & 0.833  &\textbf{0.903}& \textbf{0.897}   & 0.745      & 0.740         & 0.832         & 0.793             & 0.813             & 0.504
    \\
    \midrule
    \bottomrule
  \end{tabular}
  }
\end{table}

\begin{figure}[tbh]
\centering
\includegraphics[width=1.0\linewidth]{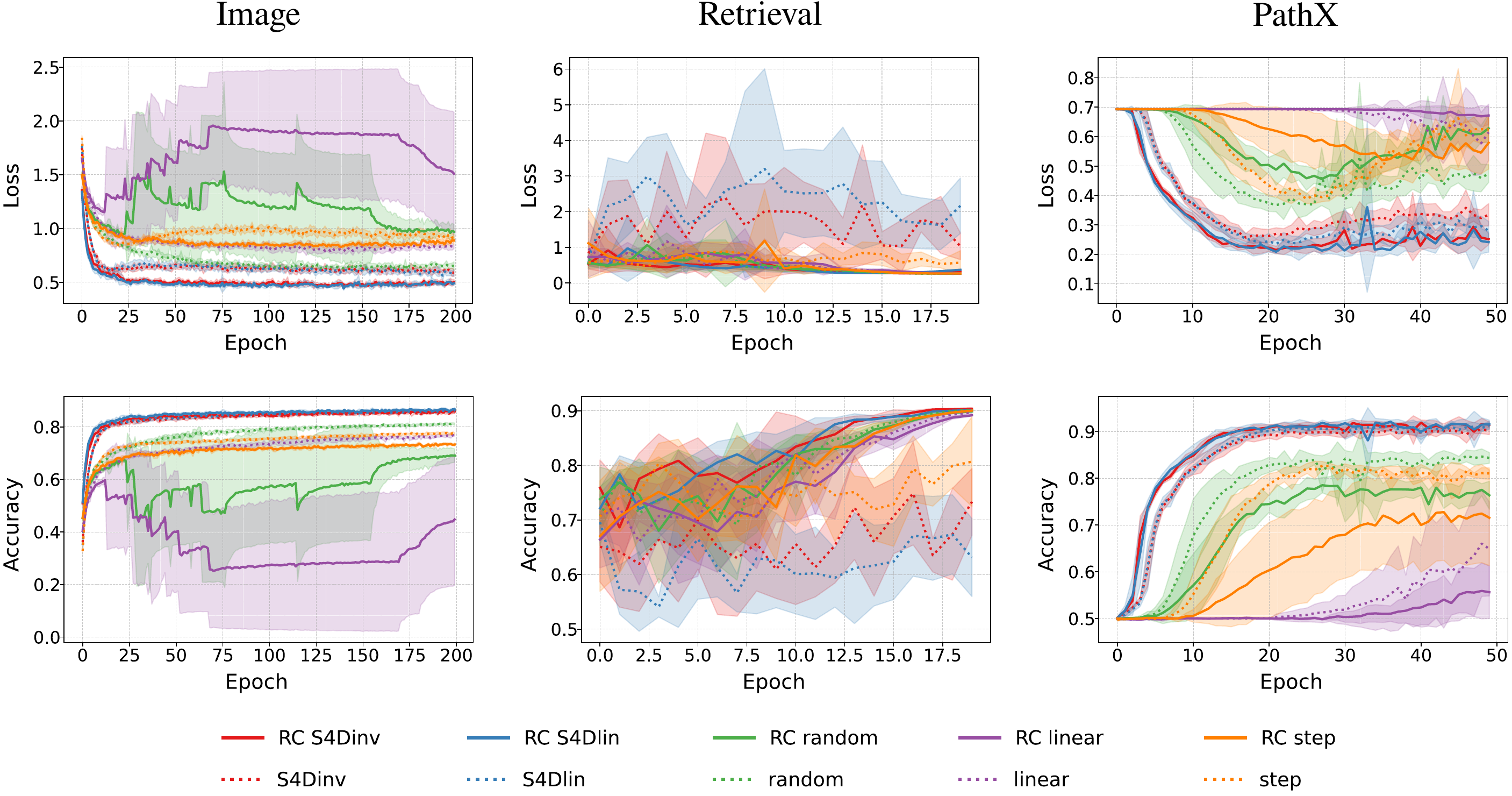}
\caption{\label{fig:acc_main}
Test loss and accuracy of three tasks.
The top (bottom) row indicates test loss (accuracy),
and each column corresponds to a different task. 
The horizontal axes represent epochs.
Lines represent averages across seeds, with shaded regions indicating standard deviations.
Solid lines represent models in the RC setting, while dotted lines indicate models with trainable eigenvalues.
Colors indicate different eigenvalue realizations.
}
\end{figure}
\begin{figure}[tbh]
\centering
\includegraphics[width=1.0\linewidth]{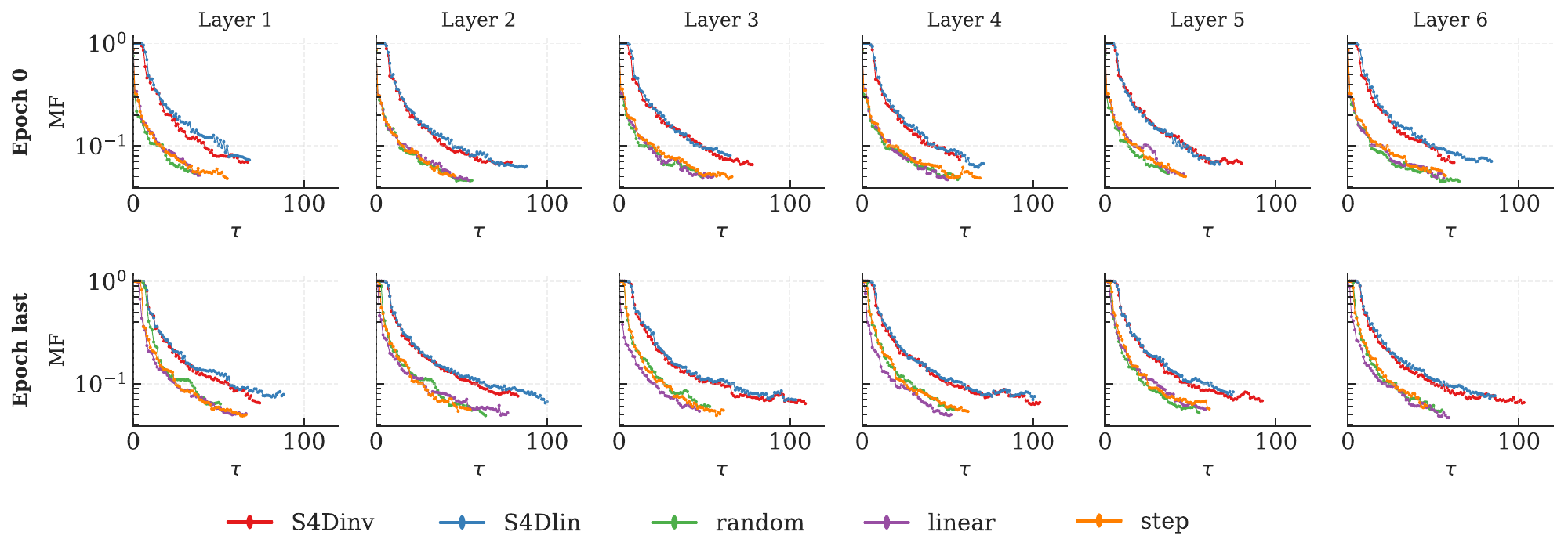}
\caption{\label{fig:S4Dinv_params_kernel}
The supremum MF of each layer before and after training.
The top (bottom) panels show MFs before (after) training.
From left to right, each column represents one layer.
The horizontal (vertical) axis shows the input delay $\tau$ (MF).
The five colors indicate the eigenvalue realizations.
The state size $N$ is $32$,
and the results are obtained by calculating Eq.~\ref{eq:original MF} with $T=1024$.
As an example, we show  the results for the Pathfinder task.
}
\end{figure}  

\begin{figure}[tbh]
\centering
\includegraphics[width=1.0\linewidth]{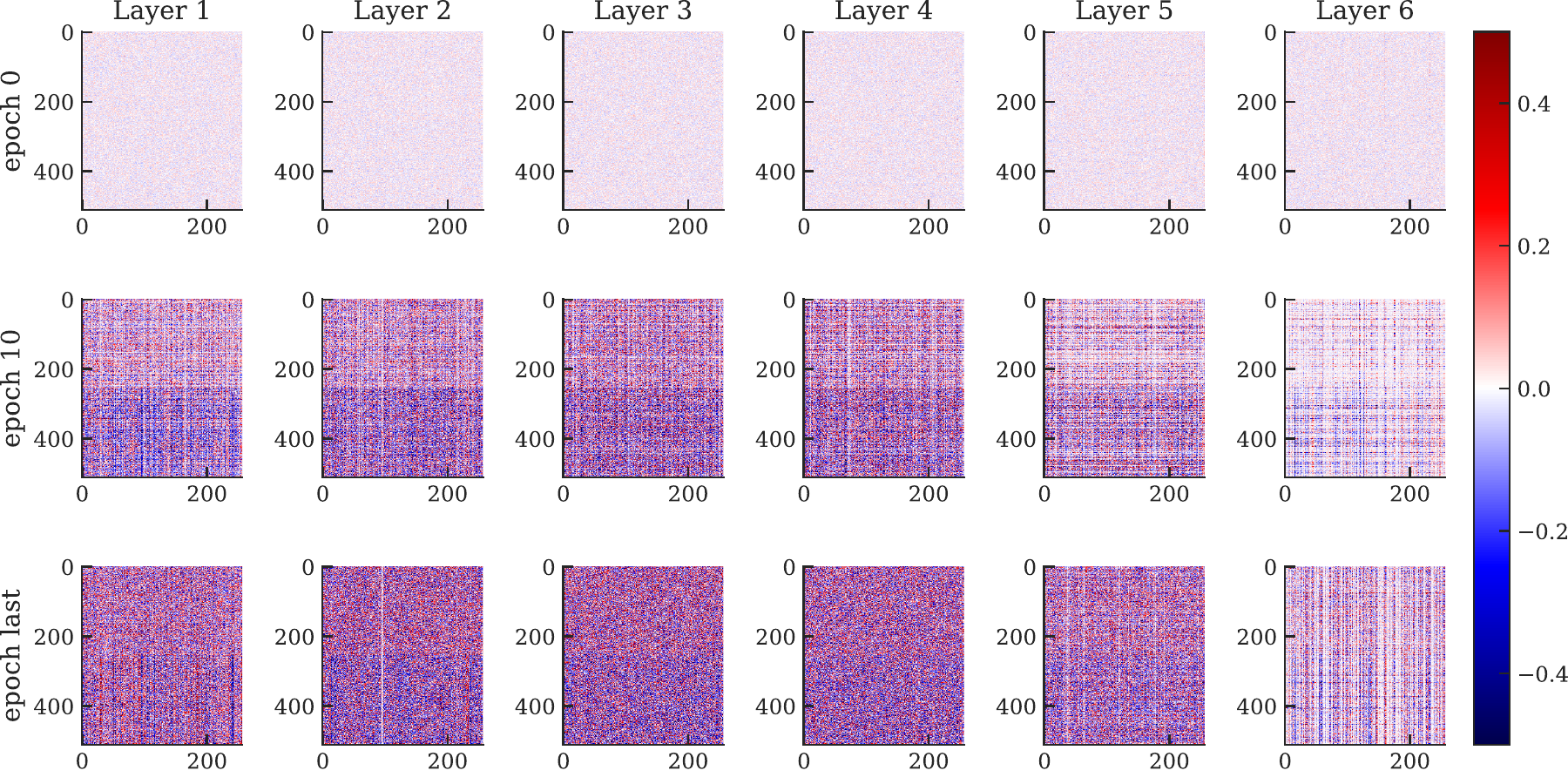}
\caption{\label{fig:S4Dinv_params_readout}
The linear readout weights $\bm{W}^{(\ell)}$ and $\bm{b}^{(\ell)}$ at each layer and epoch.
From top to bottom, the rows correspond to epochs 0, 10, and the final epoch, respectively.
From left to right, the columns show the weights of layers 1 through 6.
Following Figure~\ref{fig:S4Dinv_params_kernel}, we present an example from the Pathfinder task.
}
\end{figure}

Our experiments show that the RC setting is effective across all LRA tasks
when the eigenvalues are initialized with structured memory.
In five tasks, the RC setting with structured eigenvalues yielded higher accuracy than the setting with learnable eigenvalues
and achieved accuracies comparable to those of the S4 model.
In contrast, the model with representative eigenvalues showed poorer performance under the RC setting,
suggesting the importance of the initialization with structured eigenvalue realizations.
To examine the effects of the RC setting, we monitored how the loss and accuracy evolved over time.
Using structured eigenvalues, 
the test loss for Image, Retrieval, and PathX
converged to lower values more rapidly in the RC setting, 
and the test accuracy for Retrieval and PathX also increased notably faster (Figure~\ref{fig:acc_main}).
In addition, for the Text, Retrieval, and Image tasks,
the RC setting yielded test accuracy comparable---to or even exceeding---those obtained when eigenvalues were learned.
Moreover, the training accuracy shows a relatively modest enhancement with RC setting,
indicating better mitigation of overfitting (see Appendix~\ref{appendix:Additional results}).
For the remaining tasks, these trends were less pronounced, yet no significant contradictory outcomes were observed.
These results suggest that fixing eigenvalues during learning contributes to more efficient training.

To observe whether eigenvalue updates are necessary in detail,
we examined how the MF and readout weight of SSMs change before and after training
when learning is allowed.
As shown in Figure~\ref{fig:S4Dinv_params_kernel},
even after training, the MFs of the ``random'', ``linear'', and ``step'' eigenvalue configurations 
did not surpass those of structured eigenvalues before training.
This implies the difficulty of training eigenvalues and highlights the importance of initialization.
Focusing on each eigenvalue realization, training was able to extend the MF, suggesting the benefits of trainable eigenvalues.
However, as seen in the accuracies (Table~\ref{table:acc}),
the RC setting yielded better performance,
suggestting that learning eigenvalues may offer fewer benefits than the RC setting
if the initial eigenvalues already achieve a relatively long MF.
The reason why the extended MFs did not contribute to performance can be attributed to the factors mentioned above---namely, 
the suppression of overfitting and stabilized loss at lower values.
In Figure~\ref{fig:S4Dinv_params_readout},
the dynamics of the weights in the linear output layer are shown when eigenvalue training is allowed.
They change significantly over the course of training,
indicating that the readout learning is important even when the eigenvalues are trained.
Combined with the high scores of the RC setting,
this observation suggests that learning in SSMs may primarily progress through parameters other than eigenvalues.
\section{Discussion and Conclusion}
In this study, we theoretically demonstrated the importance of initialization and the memory requirement for successful learning,
which were not addressed in previous studies attempting to elucidate the learning mechanisms of SSMs.
In addition, those studies imposed constraints on $\bm{A}$ without sufficient theoretical justification.
For example, $\bm{A}$ was taken as random matrices~\citep{NEURIPS2020_9ac1382f},
and the eigenvectors of $\bm{A}$ were either restricted or simply ignored~\citep{smekal2024towards, proca2024learning}.
Restricting eigenvectors limits applicability to specific models,
and ignoring them confines the results to S4D.
In contrast, this study imposes no restrictions on $\bm{A}$,
thereby extending the applicability of our results to a wider range of models.
We theoretically justify that the eigenvectors of $\bm{A}$ can be disregarded,
allowing us to focus solely on the eigenvalues. 
The broader applicability of our results is discussed in Appendix~\ref{appendix:discussion}.
Furthermore, because our results rely only on gradients,
they are not limited to any specific training method and can be applied to any gradient-based optimization approach.

Using the theoretically revealed parameter update formula,
we show that the teacher information may be lost
if the SSM is not initialize with structured eigenvalues that provide sufficiently long memory.
Related research topics include the vanishing and exploding gradient problems~\citep{zucchet2024recurrent, park2023persistent, sokol2019adjoint}.
These studies focused on multilayer networks,
where gradients accumulate multiplicatively as errors propagate through the layers.
For RNNs in particular,
the problem is tied to the traditional backpropagation-through-time algorithm,
which treats the number of recurrent steps as the number of gradient accumulations.
On the other hand, the issue we highlight here arises in gradient computations within single-layer sequence-to-sequence models constructed from RNNs,
where gradient passing between recurrent steps and gradient multiplication do not occur.
Therefore, these are distinct problems with different underlying mechanisms. 
However, by using our results,
we can deduce whether vanishing and exploding gradients occur in a single layer of SSMs.
Exploding gradients occur when the maximum eigenvalue of the weight matrix exceeds one in linear SSMs,
causing the state variable time series to diverge. 
Vanishing gradients may not occur because the only case in which $G(\bm{y})=0$ (Eq.~\ref{linear: Du with MF})
is when $B\bm{V}^{\top}\bm{\theta}=\bm{V}^{\top}(\bm{V}\bm{V}^{\top})^{-1}\bm{V}\bm{\alpha}$,
indicating that learning is complete.

To validate the learning mechanism uncovered in this study,
we also introduced a novel training approach in which the eigenvalues are fixed.
As a result,
approximately $10$\% of the total parameters were removed from the set of learnable parameters.
Altough the reduction is small, we found it to be sufficient to alleviate overfitting and accelerate learning.
This effect stems from the inherent structure of SSMs and differs from mechanisms typically discussed in the broader machine learning literature.
In addition, from the perspective of designing eigenvalues, the RC setting can also be recommended.
When eigenvalues are intended to realize long memory structures, such as those derived from HiPPO,
their absolute eigenvalues tend to approach one.
Because values exceeding $1$ lead to output explosion, 
learning can make only slight adjustments to these eigenvalues,
which may support the recommendation of the RC setting (see Appendix~\ref{appendix:Additional results}).
However, one scenario in which the RC setting may not be advisable is when solving tasks that require only extremely short-term memory.
The reason for this was discussed at the end of section~\ref{sec:Training eigenvalues},
and additional analyses should take this into account.

Although we have shown that longer memory is generally preferable,
it remains an open question whether structured eigenvalues based on HiPPO (e.g., ``S4Dinv'' and ``S4Dlin'') are indeed the best choice.
The reason is that although HiPPO matrices guarantee the theoretically longest memory,
their indicator differs from the MF, which is inherently involved in the gradient.
We confirmed that more favorable MFs can be discovered under conditions where eigenvalue learning is allowed.
However, because the observed changes were modest even after convergence in accuracy and—in multiple respects—the RC setting was superior to learning,
we cannot recommend the learning approach in general.
Nonetheless, by developing an algorithm that explicitly extends the MF to achieve more desirable memory characteristics,
better models could be realized using the newly found eigenvalues within the RC setting. 
We believe this study provides a theoretical foundation for future work aimed at addressing this question.

\bibliography{iclr2026_conference}
\bibliographystyle{iclr2026_conference}

\appendix

\section{Usage of large language model}
We used a large language model (LLM) to assist non-native authors with improving the fluency and clarity of the writing. All scientific content was created and validated by the authors.
\section{Model}\label{appendix:model}
We describe the adopted model in detail here.
In principle, the setting follows prior works~\citep{guefficiently, gu2022parameterization}.
To implement the SSM on modern computers, the continuous model (Eq.~\ref{eq:continuous SSM}) is discretized using the zero-order hold (ZOH) assumption.
The input is considered to be held constant over each sampling interval of length $ \Delta$.
The system can be formulated as
\begin{align}
\begin{array}{ccc}
    \bm{x}_{k+1} = \bm{A} \bm{x}_k + \bm{B} \bm{u}_k
    & &
    \bm{y}_k = \tilde{\bm{C}} \bm{x}_k + \tilde{\bm{D}} \bm{u}_k,
\end{array}
\end{align}
where $ \bm{x}_k = \bm{x}(k \Delta) $, $ \bm{u}_k = \bm{u}(k \Delta) $,
and $ \bm{y}_k = \bm{y}(k \Delta) $ denote the sampled state, input, and output at discrete time $ k $, respectively.
The discrete system matrices $ \bm{A} $ and $ \bm{B} $ are derived from the continuous-time matrices as follows:
$\bm{A} = e^{\tilde{\bm{A}} \Delta}$ and 
$\bm{B} = \left( \int_0^{\Delta} e^{\tilde{\bm{A}} \tau} d\tau \right) \tilde{\bm{B}}$.
Following prior work on S4 and S4D,
$\Delta$ is not represented by a single fixed value within a layer but is instead assigned multiple values.
For each such $\Delta$,
independent SSMs are defined,
and within a single layer, all of these SSMs process the input in parallel.
The eigenvalues presented in Figure~\ref{fig:MF_S4} as an example are discretized using $\Delta=0.01$.

In section \ref{sec:MF of SSM} of the main text,
we used five eigenvalue realizations.
We chose them because of the following reasons.
The eigenvalues related to the HiPPO matrices called ``S4Dinv'' and ``S4Dlin'' are used
because they were proposed in previous studies~\citep{gu2022parameterization} and showed great performance.
The definition is as follows:
\begin{align}
\begin{array}{ccc}
    \text{(S4Dinv)}~~~ \lambda_m = -\frac{1}{2} + i\frac{N}{\pi}(\frac{N}{2m+1}-1)
    & &
    \text{(S4Dlin)}~~~ \lambda_m = -\frac{1}{2} + i\pi m
\end{array},
\end{align}
where $\lambda_m$ indicates the $m$th eigenvalue.
The eigenvalues named ``random'' and ``lin'' are commonly referenced examples in earlier research on the MF.
The ``step'' eigenvalues represent a case where the ``lin'' eigenvalues are extended into the complex domain
because the eigenvalues of randomly generated matrices always come in complex conjugate pairs when the matrix size $N$ is even.
When $N$ is odd, one real eigenvalue is added.
Therefore, ``step'' is more suitable than ``lin'' for comparison with random eigenvalues.

\section{MF of SSM}\label{appendix:MF of SSM}
\subsection{Evaluation}
The MF of one distinct SSM is numerically calculated through the following procedure. 
First, using Eq.~\ref{eq:discrete SSM},
we substitute the input time series $u_t$ generated by uniform randomo distribution
and compute the state values $x_t$ over a sufficiently long duration denoted as $T$.
Here, the initial $T_w$ steps are treated as a washout period.
The resulting time series of states forms the matrix $\bm{X}$ whose column represents the states at each time step,
and have the length of $T_w + T$.
Next, for each delay $\tau$, we construct a delayed input vector $U_\tau$ of length $T$. 
The MFs are obtained by substituting these $\bm{X}$ and $\bm{U}_\tau$ into the following formula:
\begin{eqnarray}
M[u_{t-\tau}] =
\frac{
    \bm{U}_{\tau}^{\top}\bm{X}(\bm{X}^{\top}\bm{X})^{-1}\bm{X}^{\top}\bm{U}_{\tau}
}{
    \bm{U}_{\tau}^{\top} \bm{U}_{\tau}
},\label{eq:normalized correlation betw UandX}
\end{eqnarray}
which is equivalent to the Eq.~\ref{eq:original MF}~\citep{dambre2012information}.
In numerical computation, due to rounding errors and noise, the MF never becomes exactly zero
even if the values are already insignificant.
To remove insignificant values, 
we used the surrogate data generated by randomly shuffling the input series in time direction.
The data preserve the statistical properties of the original data while temporal dependencies can be eliminated.
Replacing this data with the target series of MF $\bm{U}_{\tau}$,
we calculated a pseudo MF to evaluate the potential error in the target MF.
In practical, we calculated the pseudo MFs for $20$ times by generating different surrogate data
and obtain the supremum of pseudo MF by chooosing the maximum value at each $\tau$.
The target MF discards values smaller than the supremum of the pseudo MF scaled by a threshold $(=1.2)$.
Counting from the delay $\tau=0$, once elimination occurs, the MF for all subsequent $\tau$ is set to $0$.

\begin{figure}[t]
\centering
\includegraphics[width=0.7\linewidth]{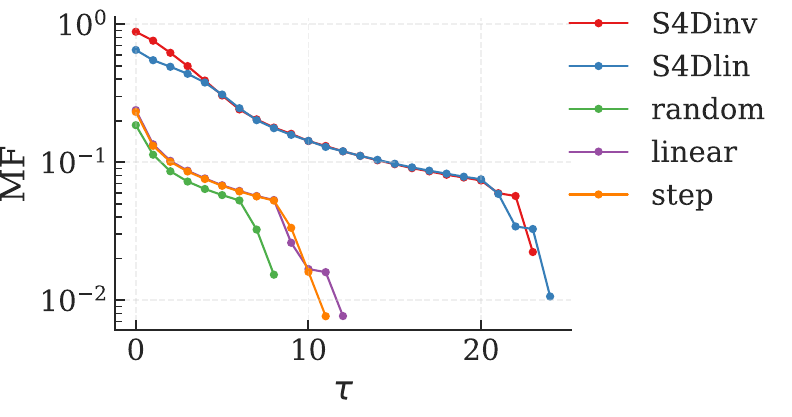}
\caption{\label{fig:MF_S4 ave}
The average MF of five eigenvalue realizations.
The horizontal axis is the delay $\tau$ of input.
The system size $N$ is $32$, and the results are obtained by calculating Eq.~\ref{eq:original MF},
where $T=1024$.
}
\end{figure}

A single SSM layer is generally composed of multiple parallel SSMs.
To evaluate an entire layer in terms of MF,
it is necessary to take into account the MFs of all constituent SSMs.
Given that, 
during forward propagation,
the input sequence of the next layer is generated from all the output sequence of the previous,
we statistically assume that all SSMs within a layer process the same input sequence.
By considering the set of MFs across all SSMs in the layer,
the maximum value at each delay $\tau$ can be interpreted as the maximum information processing capacity of the entire layer for one time series.
Accordingly, we define this supremum of MFs as the MF representing one SSM lyaer.
Here we additionally provide the average MF as supplementary information (Figure~\ref{fig:MF_S4 ave}).
As is evident, the indicator similarly shows that ``S4Dinv'' and ``S4Dlin'' have longer MFs compared to that of other eigenvalues.

\subsection{Vandermonde Matrix}
We adopted the original definition of MF (Eq.~\ref{eq:original MF})
instead of the analytical solution of MF (Eq.~\ref{eq:analytical MF}) because of the numerical unstability of the Vandermonde matrix,
which is a well studied problem~\citep{pan2016bad, SVDofStructuredMat, demmel2006accurate, drmac2015svd, batenkov2021spectral, pan2015transformations}.
In this study, the issue arises
when the MF is computed using the analytical solution of MF under the eigenvalue structures of ``S4Dinv'' and ``S4Dlin.''
During this computation,
the MF exceeds its theoretical range of $[0, 1]$,
approaching the precision limits of the machine in both positive and negative values.
The computation does not produce reasonable results,
suggesting that this analytical solution cannot be applied to these eigenvalues.
Since the MF based on the numerical computation does not exceed its theoretical range 
and the actual model behavior is expected to follow the numerical procedure,
we adopted the solution obtained numerically.

\begin{figure}[h]
\centering
\includegraphics[width=0.8\linewidth]{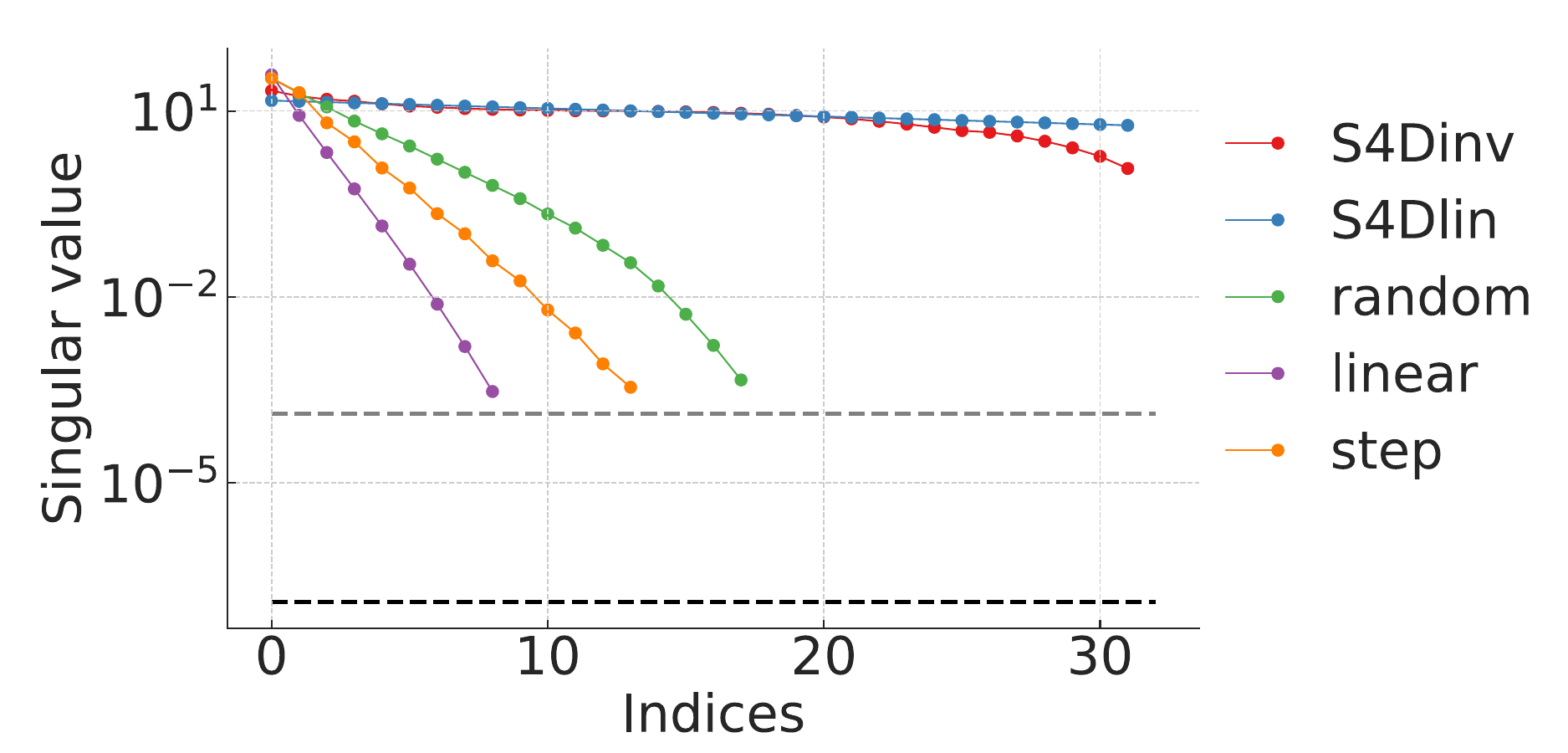}
\caption{\label{fig:H_singulars}
The singular values of matrix $\bm{V}$ calculated from five eigenvalue realizations.
The horizontal axis is an index of singular values.
The black line indicates the lower bound of computational accuracy $\epsilon$ determined by the common double precision environment.
The calculated singular value below $ \max(T, N) \times \max(\sigma)\times \epsilon$ is truncated, where $\max(\sigma)$ is the largest absolute singular values.
The maximum threshold calculated from all the realizations is depicted in the gray line.
The system size is $N=32$ and $T=1024$.
The eigenvalues used here as an example are discretized using $\Delta=0.01$.
}
\end{figure}

The issue is inherent and may arise in any computation involving $\bm{V}$.
As discussed in the main text, both MF and gradient calculation inevitably involve $\bm{V}$.
This is unavoidable when using linear RNNs, as also shown in the derivations of the original S4~\citep{guefficiently}.
However, analyzing $\bm{V}$ is still beneficial.
This is not only supported
by previous studies demonstrating the consistency between numerical computation and the analytical solution~\citep{guan2025noise},
but also by the singular values explaining the long MFs of ``S4Dinv'' and ``S4Dlin'' (Figure~\ref{fig:H_singulars}).
From the figure,
the structured eigenvalues indicates significantly more valid singular values,
explaining the results of Figure~\ref{fig:MF_S4}.

\begin{figure}[h]
\centering
\includegraphics[width=1.0\linewidth]{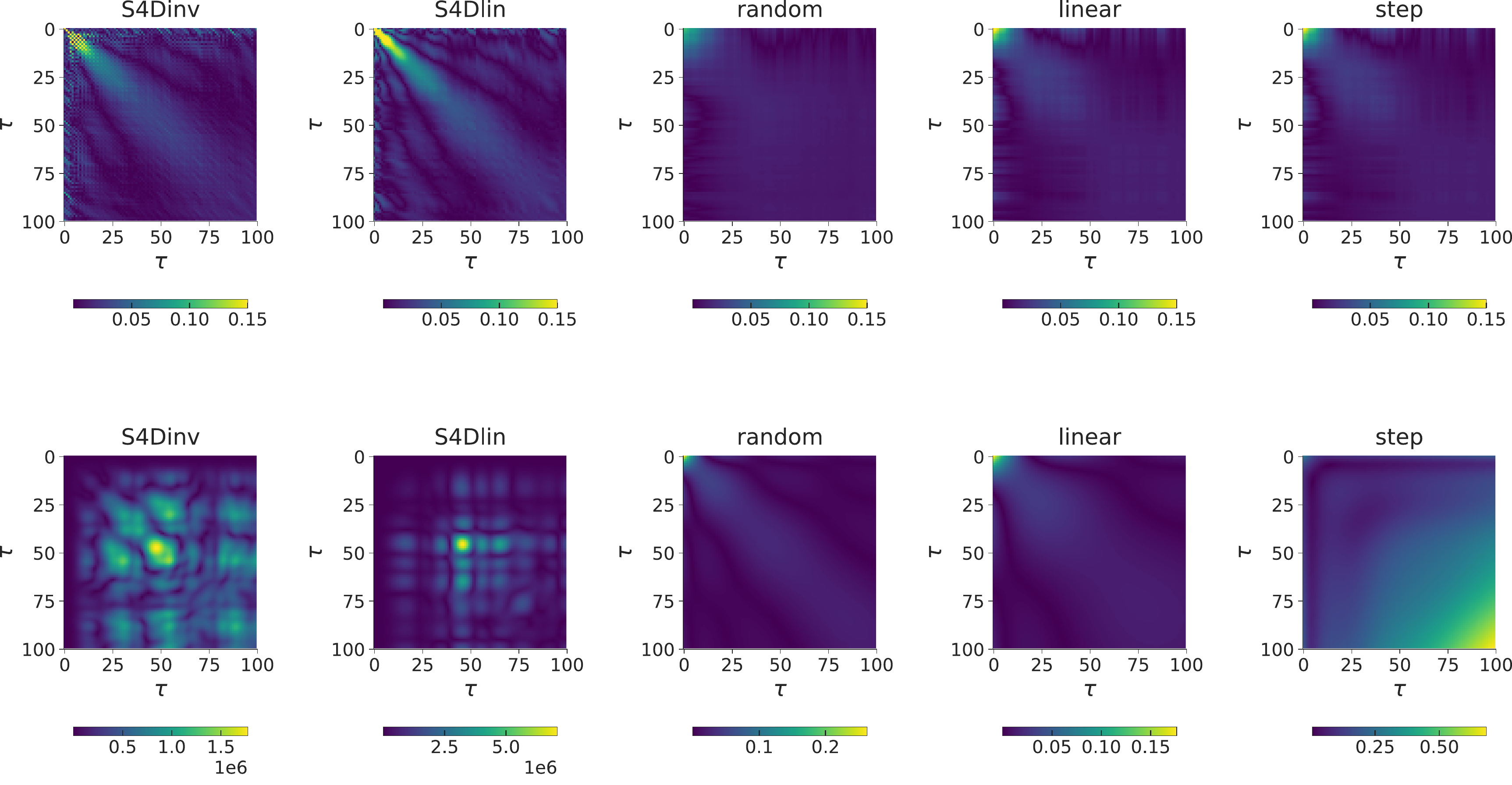}
\caption{\label{fig:VVVV}
The value of matrix $\bm{V}^{\top}(\bm{V}\bm{V}^{\top})^{-1}\bm{V}$ for five eigenvalue realizations.
The upper row shows the values following the numerical calculation,
The values in the lower row are directly computed based on the formula $\bm{V}^{\top}(\bm{V}\bm{V}^{\top})^{-1}\bm{V}$.
The range of colorbars in the upper row is consistent across all realizations
while that of the lower row depends on the result of each realizations.
Both horizontal and vertical axes indicate the delay $\tau$.
}
\end{figure}

In section~\ref{sec:Importance of initial parameters},
the cruciality of the normalized Vandermonde matrix $\bm{V}^{\top}(\bm{V}\bm{V}^{\top})^{-1}\bm{V}$ is revealed.
To illustrate that the MF can represent matrix $\bm{V}^{\top}(\bm{V}\bm{V}^{\top})^{-1}\bm{V}$ as an indicator,
we showed the values in Figure~\ref{fig:VVVV}.
Though the matrix can be computed directly,
the values obtained through this analytical solution exceed reasonable values as described in previous paragraph.
Therefore, we used a numerical method to calculate the matrix $\bm{V}^{\top}(\bm{V}\bm{V}^{\top})^{-1}\bm{V}$,
which is based on the following analytical equation:
\begin{eqnarray}
\bm{V}^{\top}(\bm{V}\bm{V}^{\top})^{-1}\bm{V}
=\frac{
    \bm{\hat{U}}^{\top}\bm{X}(\bm{X}^{\top}\bm{X})^{-1}\bm{X}^{\top}\bm{\hat{U}}
}{
    T\langle u^2\rangle
},
\end{eqnarray}
where $
\bm{\hat{U}} =
\begin{pmatrix}
    \bm{U}_{T-1} & \cdots & \bm{U}_{0}
\end{pmatrix}$ and $\langle u^2\rangle$ is the variance of $u$.
The derivation of this equation follows directly from Appendix~\ref{appendix:Parameter Update Algorithm of SSMs}.
We calculated the right hand of the equation as numerical $\bm{V}^{\top}(\bm{V}\bm{V}^{\top})^{-1}\bm{V}$.
The procedure to obtain $\bm{X}$ and $\bm{U}_\tau$ follows the method explained in the previous evaluation section

\section{The Parameter Update Algorithm of State Space Models}
\label{appendix:Parameter Update Algorithm of SSMs}
Technical details for the analysis of gradient of SSMs are provided.

\subsection{The Derivation of the Update Equation with Arbitrary Nonlinear Readout}
In this section, we analytially derive the gradient $G(\bm{y})$ of one linear SSM layer.
We begin the derivation with the formula $G(\bm{y}) = \frac{2}{T}\bm{X}^{\top}
\left[( f(\bm{X}\bm{\theta} ) - \bm{y} )
\odot f'(\bm{X}\bm{\theta} )\right]$, where $ \bm{X}\in\mathbb{R}^{T\times N} $ is a matrix whose elements consist of states series,
$T$ is the time length, $N$ is the states size,
and $f(\cdot)$ is the activation function in the readout layer. 
In structured state space sequence (S4) model, the state at time $t$ can be represented as:
\begin{eqnarray}
\bm{x}_{t}
&=&\sum_{i=1}^{t}\bm{A}^{i-1}\bm{B} u_{t-i+1}
=\sum_{i=1}^{t}\bm{P} \bm{\Lambda}^{i-1} \bm{P}^{-1}\bm{B} u_{t-i+1},
\end{eqnarray}
where initial state is $\boldsymbol{x}_0 = \boldsymbol{0}$ and $u_t$ is the input series at time $t$.
We simplified $\bm{x}_{t}$ through the eigenvalue decomposition of internal weight matrix
$\bm{A} = \bm{P} \bm{\Lambda} \bm{P}^{-1}$,
where 
$\Lambda=\text{diag}\begin{pmatrix}
\lambda_1 & \lambda_2 & \cdots& \lambda_N
\end{pmatrix}
$, and $\lambda_m$ are the $m$th eigenvalues of $\bm{A}$.
The largest absolute $\lambda_m$ is defined as the spectral radius $\rho$,
which should not exceed $1$ to ensure that the state value does not diverge.
Defining
$\bm{P} =
\begin{pmatrix}
    {\bm{p}_1} & {\bm{p}_2} & \cdots & {\bm{p}_N}
\end{pmatrix}
$
and
$\bm{P}^{-1} =
\begin{pmatrix}
    {\bm{p}_1'}^{\top}\\
    {\bm{p}_2'}^{\top}\\
    \vdots\\
    {\bm{p}_N'}^{\top}
\end{pmatrix}
$,
\begin{eqnarray}
\bm{x}_{t}
&=&\bm{P} \sum_{i=1}^{t}  u_{t-i+1} 
\text{diag}\begin{pmatrix}
\lambda_1^{i-1} & \lambda_2^{i-1} & \cdots& \lambda_N^{i-1}
\end{pmatrix}
\begin{pmatrix}
    {\bm{p}_1'}^{\top}\bm{B}\\
    {\bm{p}_2'}^{\top}\bm{B}\\
    \vdots\\
    {\bm{p}_N'}^{\top}\bm{B}
\end{pmatrix}
\\
&=&
\bm{P}
\sum_{i=1}^{t}
\begin{pmatrix}
    {\bm{p}_1'}^{\top}\bm{B}\lambda_1^{i-1}\\
    {\bm{p}_2'}^{\top}\bm{B}\lambda_2^{i-1}\\
    \vdots \\
    {\bm{p}_N'}^{\top}\bm{B}\lambda_N^{i-1}
\end{pmatrix} u_{t-i+1}.
\end{eqnarray}
Considering the $k$th delayed state $\bm{x}_{t-k}$,
the formula is given by
\begin{align*}
\bm{x}_{t-k}
&=
\bm{P}
\begin{pmatrix}
    p_{1\bm{B}}' \bm{\Lambda}_1^{\top} \bm{U}_k\\
    p_{2\bm{B}}' \bm{\Lambda}_2^{\top} \bm{U}_k\\
    \vdots \\
    p_{N\bm{B}}' \bm{\Lambda}_N^{\top} \bm{U}_k
\end{pmatrix}
=
\bm{Q}
\begin{pmatrix}
    \bm{\Lambda}_1^{\top} \bm{U}_k\\
    \bm{\Lambda}_2^{\top} \bm{U}_k\\
    \vdots\\
    \bm{\Lambda}_N^{\top} \bm{U}_k
\end{pmatrix},
\end{align*}
where
$
\bm{U}_k=
\begin{pmatrix}
    u_{t-k-(T-1)} &
    \cdots
    u_{t-k-(1)} &
    u_{t-k-(0)}
\end{pmatrix}^{\top}
$, ~
$
\bm{\Lambda}_m
=\begin{pmatrix}
    {\lambda_m}^{T-1} &
    \cdots &
    {\lambda_m}^{1} &
    {\lambda_m}^{0}
\end{pmatrix}^{\top}
$,
and 
$\bm{Q}=
\begin{pmatrix}
    {\bm{p}_1'}^{\top}\bm{B} \bm{p}_1 & {\bm{p}_2'}^{\top}\bm{B} \bm{p}_2 & \cdots & {\bm{p}_N'}^{\top}\bm{B} \bm{p}_N
\end{pmatrix}
$.
As a result, $\bm{X}$ is described as follows:
\begin{eqnarray}
\bm{X}^{\top}
=
\bm{Q}
\left(
\begin{array}{cccc}
     \bm{\Lambda}_1^{\top} \bm{U}_{T-1}\\
    \bm{\Lambda}_2^{\top} \bm{U}_{T-1}\\
    \vdots\\
    \bm{\Lambda}_N^{\top} \bm{U}_{T-1}
\end{array}
\cdots
\begin{array}{cccc}
    \bm{\Lambda}_1^{\top} \bm{U}_1& \bm{\Lambda}_1^{\top} \bm{U}_0\\
    \bm{\Lambda}_2^{\top} \bm{U}_1& \bm{\Lambda}_2^{\top} \bm{U}_0\\
    \vdots & \vdots\\
    \bm{\Lambda}_N^{\top} \bm{U}_1& \bm{\Lambda}_N^{\top} \bm{U}_0
\end{array}
\right)
=
\bm{Q}
\left(
\begin{array}{cccc}
    \bm{\Lambda}_1^{\top} \\
    \bm{\Lambda}_2^{\top} \\
    \vdots\\
    \bm{\Lambda}_N^{\top}
\end{array}
\right)
\begin{pmatrix}
    \bm{U}_{T-1} & \cdots & \bm{U}_1 & \bm{U}_0
\end{pmatrix}.
\end{eqnarray}

Here, we go back the formula of the gradient:
\begin{eqnarray}
G(\bm{y})
&=&
\frac{2}{B}\bm{X}^{\top}
\left[
    f(\bm{X}\bm{\theta} )\odot f'(\bm{X}\bm{\theta} )
    - \bar{\bm{y}}
\right]
\\
&=&
\frac{2}{B}\bm{X}^{\top}
\left[
    f(\bm{X}\bm{\theta} )\odot f'(\bm{X}\bm{\theta} )
    - \bm{X}(\bm{X}^{\top}\bm{X})^{-1}\bm{X}^\top \bar{\bm{y}}
\right],
\label{eq:nonlin_activation_grad}
\end{eqnarray}
where $\bar{\bm{y}} = \bm{y} \odot f'(\bm{X}\bm{\theta})$ represents the label modified by the derivative of the activation function.
The covariance matrix $\bm{X}^{\top}\bm{X}$ has the following form:
\begin{eqnarray}
\bm{X}^{\top}\bm{X}
=
\bm{Q}
\bm{V}
\begin{pmatrix}
    \bm{U}_{T-1} & \cdots & \bm{U}_0
\end{pmatrix}
\begin{pmatrix}
    \bm{U}_{T-1} & \cdots & \bm{U}_0
\end{pmatrix}^{\top}
\bm{V}^{\top}
\bm{Q}^{\top},
\end{eqnarray}
where $\bm{V}=
\begin{pmatrix}
    {\lambda_1}^{T-1} & {\lambda_1}^{T-2} & \cdots & {\lambda_1}^{1} & 1\\
    {\lambda_2}^{T-1} & {\lambda_2}^{T-2} & \cdots & {\lambda_2}^{1} & 1\\
    \vdots\\
    {\lambda_N}^{T-1} & {\lambda_N}^{T-2} & \cdots & {\lambda_N}^{1} & 1\\
\end{pmatrix}
$.
Similar to the MF, to evaluate the inherent statistical structure,
we assume that the input $u_{t}$ is an independent and identically distributed random variable,
suggesting that delayed time series $\bm{U}_k$ are orthogonal to each other.
Using $\begin{pmatrix}
    \bm{U}_{T-1} & \cdots & \bm{U}_0
\end{pmatrix}
=
\begin{pmatrix}
    \bm{U}_{T-1} & \cdots & \bm{U}_0
\end{pmatrix}^{\top}$,
the covariance matrix $\bm{X}^{\top}\bm{X}$ reduces to the following form:
\begin{eqnarray}
\bm{X}^{\top}\bm{X}
=
\bm{Q}
\bm{V}
\begin{pmatrix}
    {\bm{U}_{T-1}}^{\top} \\
    \vdots \\
    {\bm{U}_{0}}^{\top}
\end{pmatrix}
\begin{pmatrix}
    \bm{U}_{T-1} & \cdots & \bm{U}_0
\end{pmatrix}
\bm{V}^{\top}
\bm{Q}^{\top}
=T\langle u^2\rangle
\bm{Q} \bm{V}\bm{V}^{\top}\bm{Q}^{\top},
\end{eqnarray}
where $\langle u^2\rangle$ is the variance of $u$.
Since
\begin{eqnarray}
\bm{X}(\bm{X}^{\top}\bm{X})^{-1}\bm{X}^\top
=
(T\langle u^2\rangle)^{-1}
\begin{pmatrix}
    \bm{U}_{T-1} & \cdots & \bm{U}_0
\end{pmatrix}^{\top}
\bm{V}^{\top} (\bm{V}\bm{V}^{\top})^{-1} \bm{V}
\begin{pmatrix}
    \bm{U}_{T-1} & \cdots & \bm{U}_0
\end{pmatrix},
\end{eqnarray}
we could observe that the gradient of SSM (Eq. \ref{eq:nonlin_activation_grad})
is determined by the matrix $\bm{V}^{\top} (\bm{V}\bm{V}^{\top})^{-1} \bm{V}$.
As a result, the gradient $G(\bm{y})$ is described by
\begin{eqnarray}
G(\bm{y})
&=&
\frac{2}{B}\bm{X}^{\top}
\left[
    f(\bm{X}\bm{\theta} )\odot f'(\bm{X}\bm{\theta} )
    -
    \frac{1}{T\langle u^2\rangle}
    \bm{\hat{U}}^{\top}
    \bm{V}^{\top}(\bm{V}\bm{V}^{\top})^{-1}\bm{V}
    \bm{\hat{U}}^\top \bar{\bm{y}}
\right],
\end{eqnarray}
where $
\bm{\hat{U}} =
\begin{pmatrix}
    \bm{U}_{T-1} & \cdots & \bm{U}_{0}
\end{pmatrix}$.
Using the fact that linear SSM only performs linear information processing,
all possible output of the SSM can be described as
the linear combination of past inputs series $
\bar{\bm{y}} = \sum_{k=0}^{T-1} \bar{\alpha}_k\bm{U}_k
=\bm{\hat{U}} \bar{\bm{\alpha}}
$,
where $\{\bar{\alpha}_{\tau}\}_{\tau=0}^{T-1}$ is a certain series characterizing $\bar{\bm{y}}$
and 
$\bar{\bm{\alpha}} = \begin{pmatrix}
    \bar{\alpha}_{T-1} & \cdots & \bar{\alpha}_{0}
\end{pmatrix}^\top
$.
By substituting this $\bar{\bm{y}}$
into $G(\bm{y})$,
the gradient is computed as
\begin{eqnarray}
G(\bm{y})
&=&
\frac{2}{B}\bm{X}^{\top}
\left[
    f(\bm{X}\bm{\theta} )\odot f'(\bm{X}\bm{\theta} )
    -
    \frac{1}{T\langle u^2\rangle}
    \bm{\hat{U}}^{\top}
    \bm{V}^{\top}(\bm{V}\bm{V}^{\top})^{-1}\bm{V}
    \bm{\hat{U}}^\top \bm{\hat{U}} \bar{\bm{\alpha}}
\right]
\\
&=&
\frac{2}{B}\bm{X}^{\top}
\left[
    f(\bm{X}\bm{\theta} )\odot f'(\bm{X}\bm{\theta} )
    -
    \bm{\hat{U}}^{\top}
    \bm{V}^{\top}(\bm{V}\bm{V}^{\top})^{-1}\bm{V}
    \bar{\bm{\alpha}}
\right].
\end{eqnarray}

\subsection{The Update Equation with Linear Activation Function}
we consider the linear activation case,
To reveal how information processing of linear SSMs affect the gradient,
we focus on the gradient before the nonllinear activation layer,
where gradient is described by
\begin{eqnarray}
G(\bm{y})
&=&
\frac{2}{T} \bm{X}^{\top}(\bm{X}\bm{\theta} - \bm{y}).
\end{eqnarray}
Following the main text, we consider a typical task, where $\bm{y}$ is the delayed input series $\bm{U}_{\tau}\in\mathbb{R}^{T}$,
\begin{eqnarray}
G(\bm{U}_{\tau})
&=&
\frac{2}{T} (\bm{X}^{\top}\bm{X}\bm{\theta} - \bm{X}^{\top}\bm{U}_{\tau}),
\end{eqnarray}
where $\tau$ is the delay.
Because
\begin{eqnarray}
\bm{X}^{\top}\bm{U}_{\tau}
=\bm{Q}
\bm{V}
\begin{pmatrix}
    {\bm{U}_{T-1}}^{\top} \\
    \vdots \\
    {\bm{U}_{0}}^{\top}
\end{pmatrix}\bm{U}_{\tau}
=\bm{Q}
\bm{V}
\begin{pmatrix}
    0\\
    \vdots \\
    0\\
    \bm{U}_{\tau}^{\top}\bm{U}_{\tau} \\
    0\\
    \vdots\\
    0
\end{pmatrix}
=T\langle u^2\rangle\bm{Q}
\bm{V}_{\tau},
\end{eqnarray}
where $\bm{V}_{\tau}=
\begin{pmatrix}
    {\lambda_1}^{\tau} &
    {\lambda_2}^{\tau} &
    \cdots &
    {\lambda_N}^{\tau}
\end{pmatrix}^{\top}$,
combining the result of nonlinear activation case,
we obtain
\begin{eqnarray}
G(\bm{U}_{\tau})
=
\frac{2}{T}
(
    T\langle u^2\rangle \bm{Q} \bm{V}\bm{V}^{\top}\bm{Q}^{\top}\bm{\theta}
    - T\langle u^2\rangle\bm{Q} \bm{V}_{\tau}
)
=
2\langle u^2\rangle
(\bm{Q} \bm{V}\bm{V}^{\top}\bm{Q}^{\top}\bm{\theta}
    -\bm{Q} \bm{V}_{\tau}
),
\bm{}
\end{eqnarray}
Following the framework of MF introduced in the main text, we simplify S4 to S4D,
indicating that $\bm{P}=\bm{E}$ and
$\bm{Q} = 
\text{diag}
\begin{pmatrix}
\bm{B}
\end{pmatrix}$.
Using $\bm{B}=B\bm{1}$, we can compute the gradient as
\begin{eqnarray}
G(\bm{U}_{\tau})
&=&
2\langle u^2\rangle\bm{V}\bm{V}^{\top}(
    B^2\bm{\theta}
    - B (\bm{V}\bm{V}^{\top})^{-1}\bm{V}_{\tau}
)\\
&=&
2\langle u^2\rangle B\bm{V} (
    B\bm{V}^{\top}\bm{\theta}
    - \bm{V}^{\top}(\bm{V}\bm{V}^{\top})^{-1}\bm{V}_{\tau}
).
\end{eqnarray}

All the possible gradient can be obtained by substituting the output $
\bm{y}=\sum_{k=0}^{T-1}\alpha_k\bm{U}_k
$ because the linear combination of the past inputs series fully describes the output of the SSM,
where $\{\alpha_{\tau}\}_{\tau=0}^{T-1}$ is a certain series characterizing $\bm{y}$.
\begin{eqnarray}
G(\bm{y})
&=&
2\langle u^2\rangle\bm{V}\bm{V}^{\top}(
    B^2\bm{\theta}
    - B \sum_{k=0}^{T-1}\alpha_k(\bm{V}\bm{V}^{\top})^{-1}\bm{V}_{k}
)
\\
&=&
2\langle u^2\rangle B\bm{V}(
    B\bm{V}^{\top}\bm{\theta}
    - \bm{V}^{\top}(\bm{V}\bm{V}^{\top})^{-1}\bm{V}\bm{\alpha}
).
\end{eqnarray}

\subsection{The importance of MF}\label{appendix:importance of MF}

For example, we consider SGD.
The terminal parameter $\bm{\theta}$ can be analytically derived.
$\bm{\theta}$ is obtained by solving the equation
\begin{eqnarray}
\bm{\theta}
&=&\bm{\theta} - \eta_t 2\langle u^2\rangle (
    B^2\bm{V}\bm{V}^{\top}\bm{\theta}
    - B\bm{V}\bm{\alpha}
    ).
\end{eqnarray}
As a result, we obtain $\bm{\theta}$ and the output $\bm{X}\bm{\theta}$ as follows:
\begin{eqnarray}
\bm{\theta}
&=&
\frac{1}{B} 
(\bm{V}\bm{V}^{\top})^{-1}\bm{V}\bm{\alpha},
\\
\bm{X}\bm{\theta}
&=&
\begin{pmatrix}
    \bm{U}_{T-1} & \cdots & \bm{U}_{0}
\end{pmatrix}^{\top}
\bm{V}^{\top}(\bm{V}\bm{V}^{\top})^{-1}\bm{V}
\bm{\alpha},
\label{eq:solution SGD}
\end{eqnarray}
where matrix $\bm{V}^{\top}
(\bm{V}\bm{V}^{\top})^{-1}\bm{V}$ appears again,
suggesting that MF is inherent in the update algorithm.
Actually, this is a reasonable result, as both SGD and MF are derived using the MSE as their loss function.
The result obtained here proves the equivalence of the two optimization strategies in a statistical view.

The role of MF can be further revealed from Eqs.~\ref{linear: Du with MF} and \ref{eq:solution SGD},
especially in the following part:
\begin{eqnarray}
\bm{V}^{\top}(\bm{V}\bm{V}^{\top})^{-1}\bm{V}
\bm{\alpha},
\end{eqnarray}
as $\bm{\alpha}$ is itself the MF of $\bm{y}$.
This follows directly from Eq.~\ref{eq:normalized correlation betw UandX}
by replacing $\bm{X}$ with $\bm{y}$
when $u$ is regarded as the source of $\bm{y}$.
This is in fact straightforward given the definition of $\alpha$.
The result proposes that
the gradient of SSM is statistically characterized by the product of the two MFs:
MF of the SSM itself and MF of the teacher information $\bm{y}$.

\section{Experimental Details}~\label{appendix:Experimental Details}
In this section, we provide the detailed methods and additional results for the experiments.

\subsection{Task}
The task of LRA benchmark includes long listops (Listops), byte-level text classification (Text),
byte-level document retrieval (Retrieval), image classification on sequences of pixels (Image),
a task to determine from pixel sequences whether a valid path exists between two points (Pathfinder),
and another more difficult Pathfinder task with longer sequence (PathX).

\subsection{Hyperparameters and Environment}
The model are constructed following the official Git repository\footnote{\url{https://github.com/state-spaces/s4/tree/simple}}
provided by the authors of S4~\citep{guefficiently}.
According to the result that S4 and S4D are equivalent,
we focus on the S4D model which has lower computational cost.
The hyperparameters are described in the Table~\ref{table:params}.
\texttt{n\_layers}, \texttt{d\_model}, and \texttt{n\_ssm}
denote the number of layers, the number of SSMs per layer, and the number of distinct matrices $\bm{A}$, respectively.
\texttt{lr}, \texttt{lr\_layer}, and \texttt{lr\_dt},
indicate the learning rate for the entire model, the learning rate for each layer,
and the learning rate for $\Delta$, respectively.
\texttt{dt\_min} and \texttt{dt\_max} correspond to the minimum and maximum values of $\Delta$, respectively.
In the implementation, $N$ is defined by \texttt{d\_state}$/2$,
which produces $N=32$ as specified in the main text.
The activation fucntion is defined as $f=$ GELU.
In practical, a bidirectional operation was introduced.
With this operation, the SSM kernel outputs two time series:
the sequence corresponding to the original ordering of timesteps,
and an additional time-reversed sequence.
As a result, the linear readout, $\bm{W}^{(\ell)}$ and $\bm{b}^{(\ell)}$,
receives an input sequence that is twice the length of the layer input time series.
In addition, after the linear readout layer, post activation $g=$ GLU was introduced.
Cross-entropy is employed as the loss function in the final layer.
Our results confirm that the findings are consistent with the theoretical outcomes obtained using MSE.

The computations were performed using two NVIDIA A100 Tensor Core GPU [40GB] and CPU with 4 threads.
The maximum memory usage of the CPU was approximately 20 GB to run the training.
Excluding, PathX, each training and evaluation session was completed within approximately five hours.
For PathX, it took about twenty hours.

\begin{table}[htbp]
\centering
\caption{Hyperparameters of the model}
\label{table:params}
\begin{tabular}{@{}llcccccc@{}}
\toprule
parameter               & Listops   & Text      & Retrieval & Image     &Pathfinder &PathX\\
\midrule
\texttt{n\_layers}      & 8         & 6         & 6         & 6         & 6         & 6\\
\texttt{d\_model}       & 128       & 256       & 256       & 512       & 256       & 256\\
\texttt{d\_state}       & 64        & 64        & 64        & 64        & 64        & 64\\
\texttt{n\_ssm}         & 128       & 256       & 256       & 2         & 256       & 256\\
\texttt{lr}             & 0.01      & 0.01      & 0.01      & 0.01      & 0.004     & 0.0005\\
\texttt{lr\_layer}      & 0.001     & 0.001     & 0.001     & 0.001     & 0.001     & 0.0005\\
\texttt{lr\_dt}         & 0.01      & 0.01      & 0.01      & 0.001     & 0.001     & 0.0005\\
\texttt{dt\_min}        & 0.001     & 0.001     & 0.001     & 0.001     & 0.001     & 0.0001\\
\texttt{dt\_max}        & 0.1       & 0.1       & 0.1       & 0.1       & 0.1       & 0.01\\
\texttt{dropout}        & 0.        & 0.        & 0.        & 0.1       & 0.        & 0.\\
\texttt{weight\_decay}  & 0.05      & 0.05      & 0.05      & 0.05      & 0.03      & 0.05\\
\texttt{batch\_size}    & 50        & 16        & 32        & 50        & 64        & 16\\
\texttt{max\_eopch}     & 40        & 32        & 20        & 200       & 200       & 50\\
\bottomrule
\end{tabular}
\end{table}

\subsection{Additional results}~\label{appendix:Additional results}
\begin{figure}[tbh]
\centering
\includegraphics[width=1.0\linewidth]{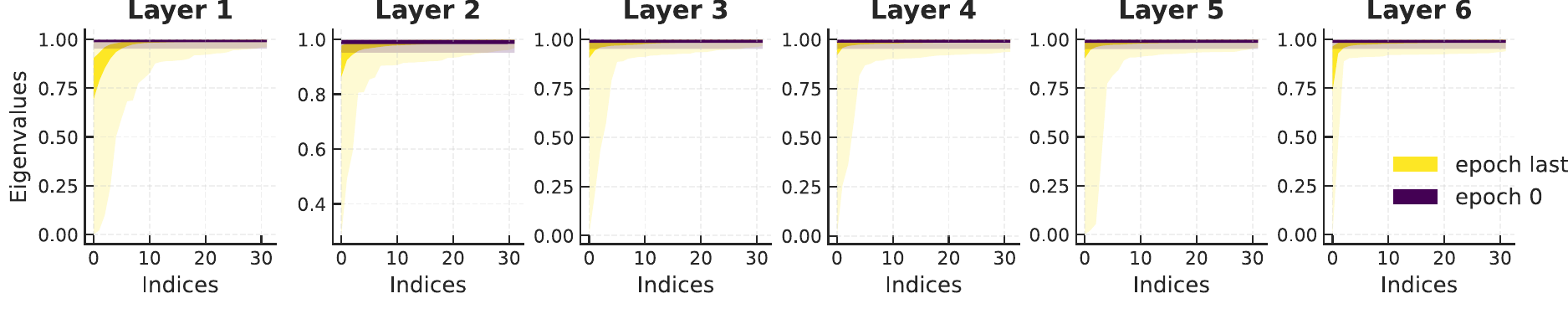}
\caption{ \label{fig:eigs_s4dinv_pathfinder}
The eigenvalues of five realizations before and after training.
The horizontal axis is the index.
Color indicates the epoch.
The dark shading represents the interquartile range between Q1 and Q3,
while the light shading indicates the range from the minimum to the maximum.
The system size $N$ is $32$.
As an exmaple, the case of Pathfinder task is shown.
}
\end{figure}

\begin{figure}[tbh]
\centering
\includegraphics[width=1.0\linewidth]{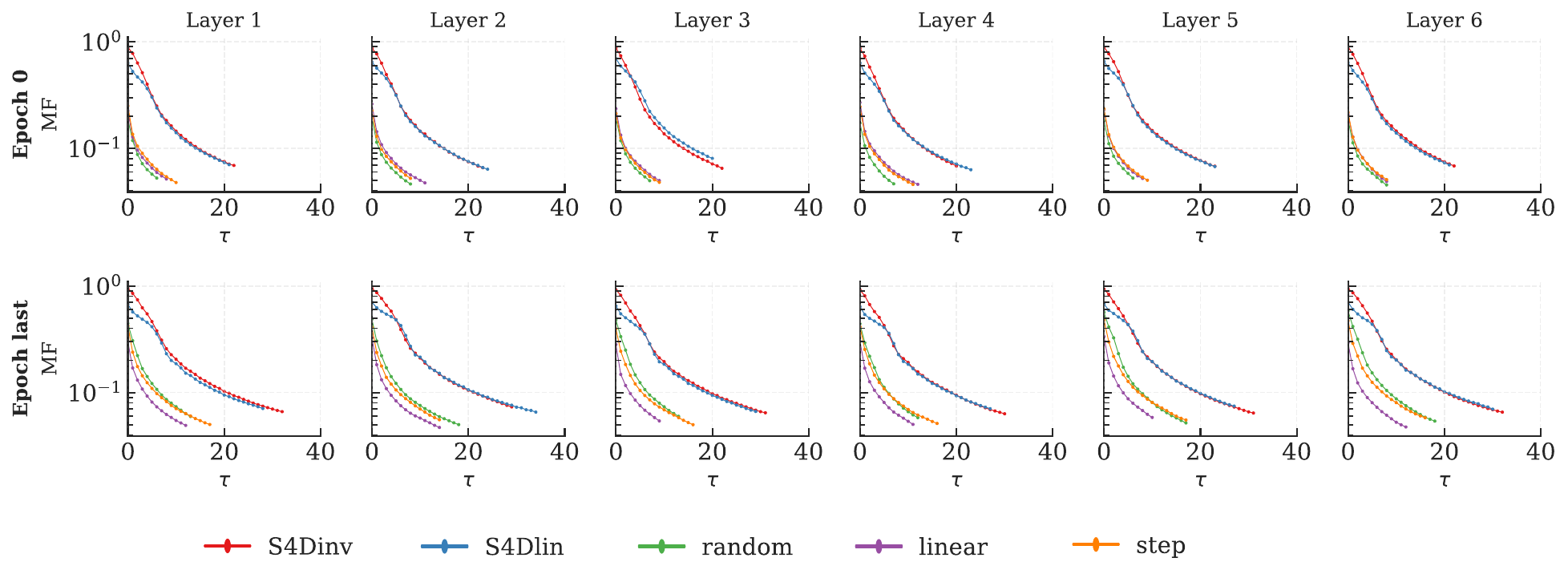}
\caption{\label{fig:MF_ave_eT1024}
The MFs of each layer before and after training.
The upper (lower) panels show MFs before (after) training.
From left to right, the column indicates the each layer.
The horizontal (vertical) axes are delay $\tau$ of input (MF).
The five colors indicates the eigenvalue realizations.
The state size $N$ is $32$, and the results are obtained by calculating Eq.~\ref{eq:original MF},
where $T=1024$.
As an exmaple, we show the case of Pathfinder task.
}
\end{figure}

In Figure~\ref{fig:S4Dinv_params_kernel}, we showed the supremum MF of one SSM layer
when the training of eigenvalues is allowed.
To demonstrate completeness, we additionally provide the average MF before and after training (Figure~\ref{fig:MF_ave_eT1024}).
As with the supremum MF,
the MF for the structured eigenvalues ``S4Dinv'' and ``S4Dlin'' exhibits longer memory compared to other eigenvalues.
To reveal the reason of the modest change in the MFs,
we show the eigenvalues before and after training (Figure~\ref{fig:eigs_s4dinv_pathfinder}).
From the initial setting, the absolute eigenvalues are almost $1$,
and this characteristic is maintained even after training.

\clearpage
Furthermore, the loss and accuracy during the training and test phase are provided as supplementary materials.


\begin{figure}[h]
\centering
\includegraphics[width=1.0\linewidth]{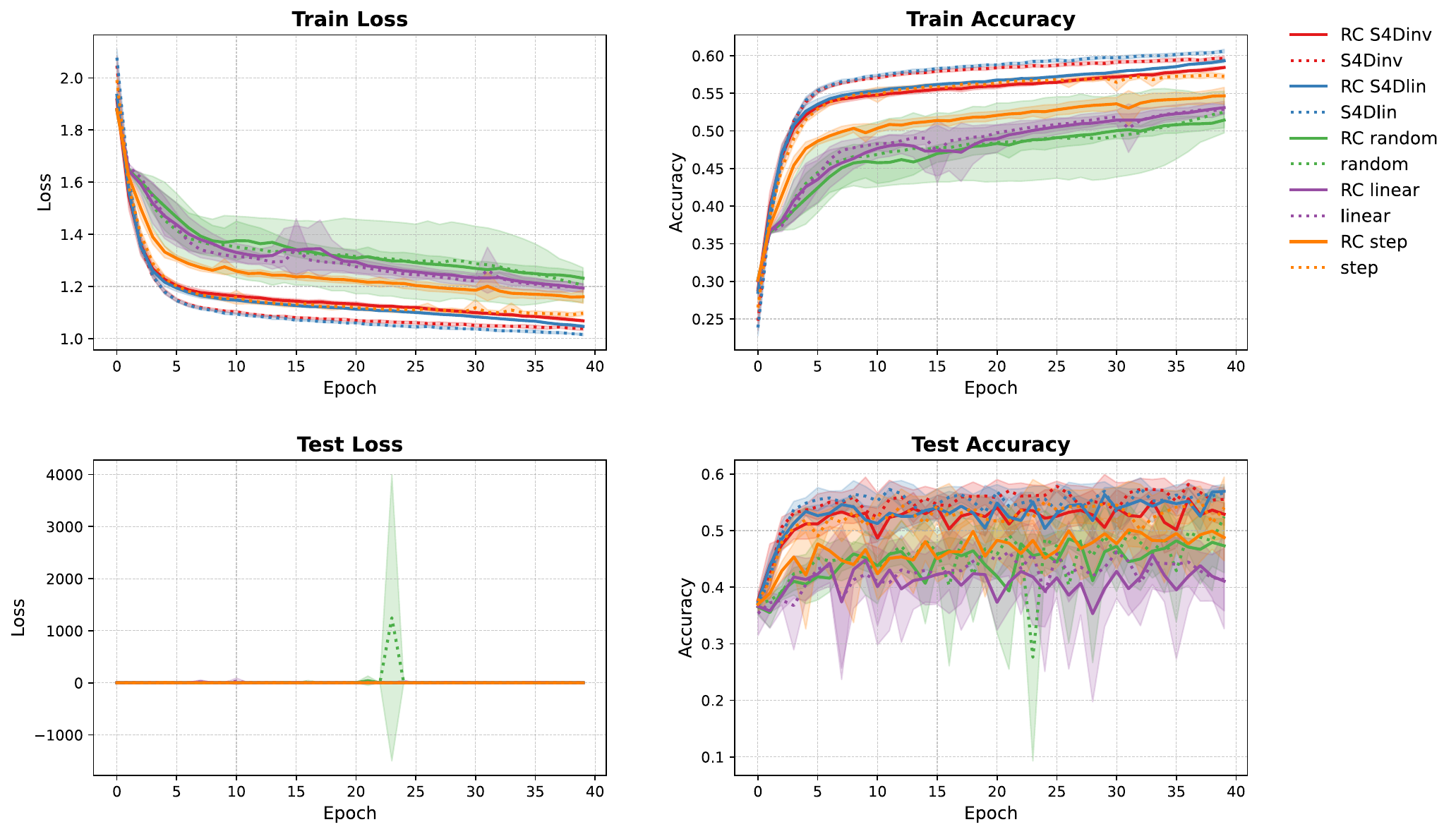}
\caption{\label{fig:acc_listops}
Test loss and accuracy of during the learning of Listops task.
The upper (lower) row indicates the metrics of train (test).
The column corresponds to each metric.
The horizontal axes are epochs.
The colors indicate eigenvalue realizations.
The solid lines are the model under RC setting
while the dotted lines are the model whose eigenvalues are allowed to learn.
}
\end{figure}

\begin{figure}[h]
\centering
\includegraphics[width=1.0\linewidth]{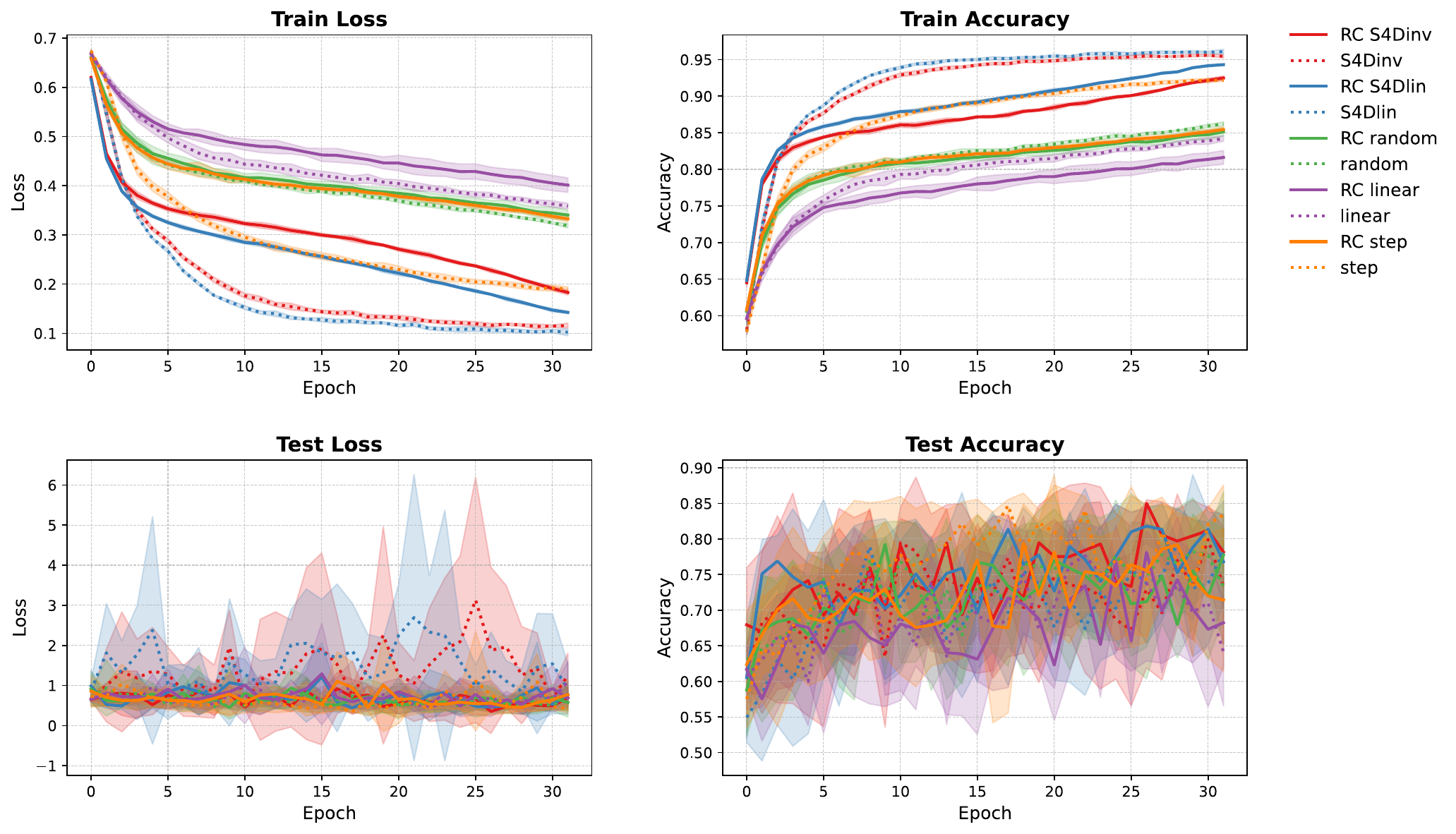}
\caption{\label{fig:acc_text}
Test loss and accuracy of during the learning of Text task.
The upper (lower) row indicates the metrics of train (test).
The column corresponds to each metric.
The horizontal axes are epochs.
The colors indicate eigenvalue realizations.
The solid lines are the model under RC setting
while the dotted lines are the model whose eigenvalues are allowed to learn.
}
\end{figure}

\begin{figure}[h]
\centering
\includegraphics[width=1.0\linewidth]{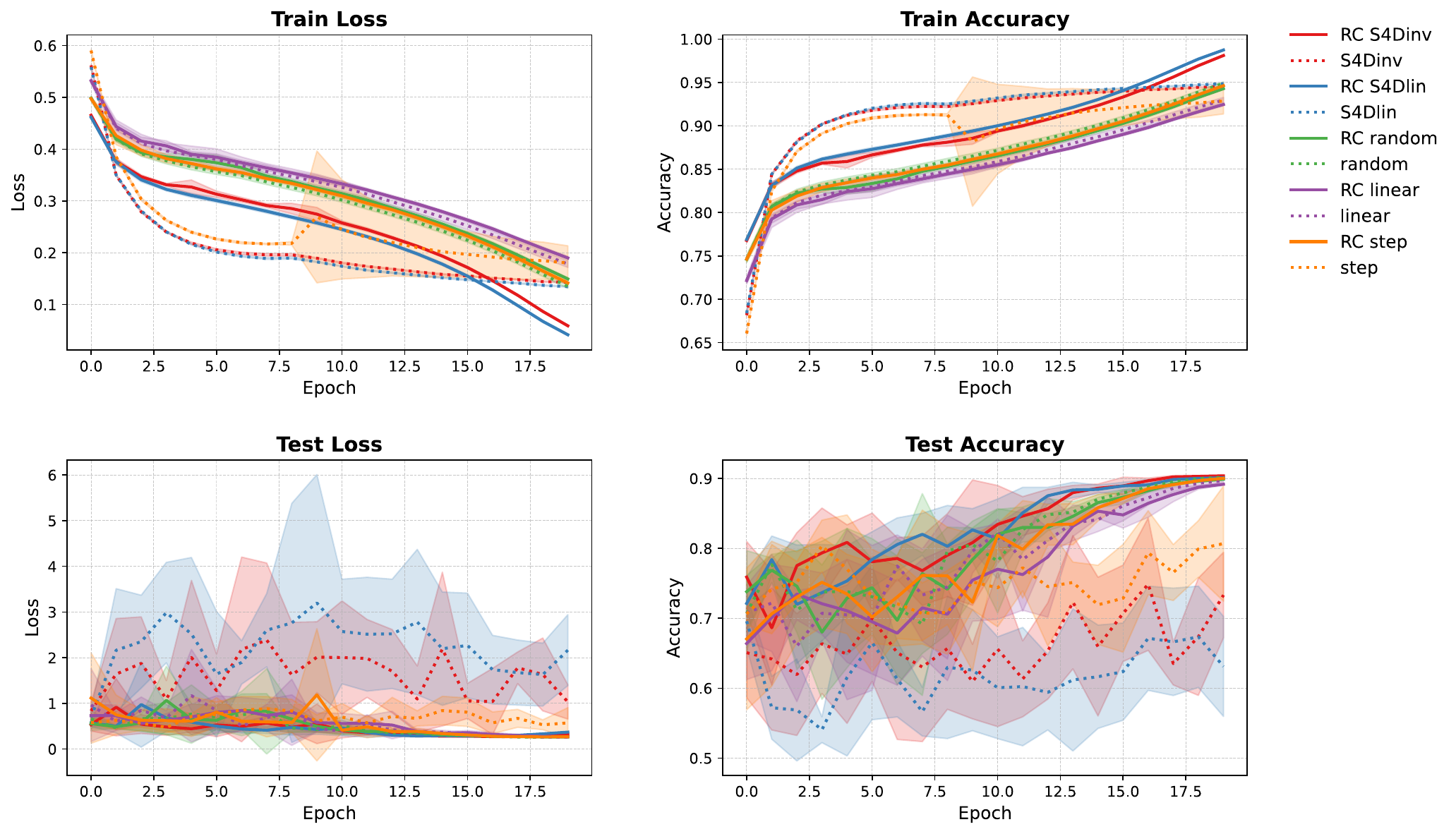}
\caption{\label{fig:acc_retrieval}
Test loss and accuracy of during the learning of Retrieval task.
The upper (lower) row indicates the metrics of train (test).
The column corresponds to each metric.
The horizontal axes are epochs.
The colors indicate eigenvalue realizations.
The solid lines are the model under RC setting
while the dotted lines are the model whose eigenvalues are allowed to learn.
}
\end{figure}

\begin{figure}[h]
\centering
\includegraphics[width=1.0\linewidth]{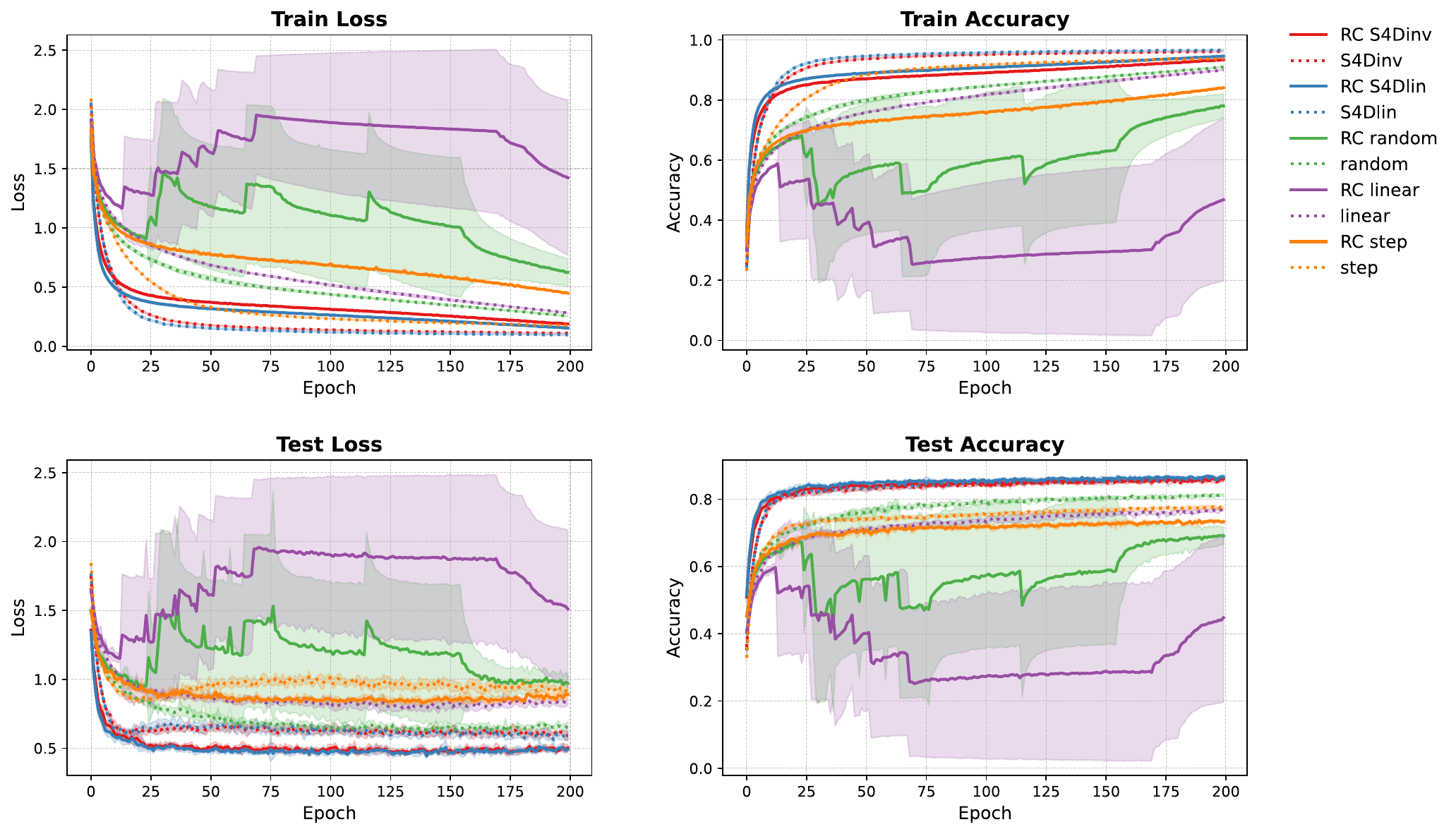}
\caption{\label{fig:acc_image}
Test loss and accuracy of during the learning of Image task.
The upper (lower) row indicates the metrics of train (test).
The column corresponds to each metric.
The horizontal axes are epochs.
The colors indicate eigenvalue realizations.
The solid lines are the model under RC setting
while the dotted lines are the model whose eigenvalues are allowed to learn.
}
\end{figure}

\begin{figure}[h]
\centering
\includegraphics[width=1.0\linewidth]{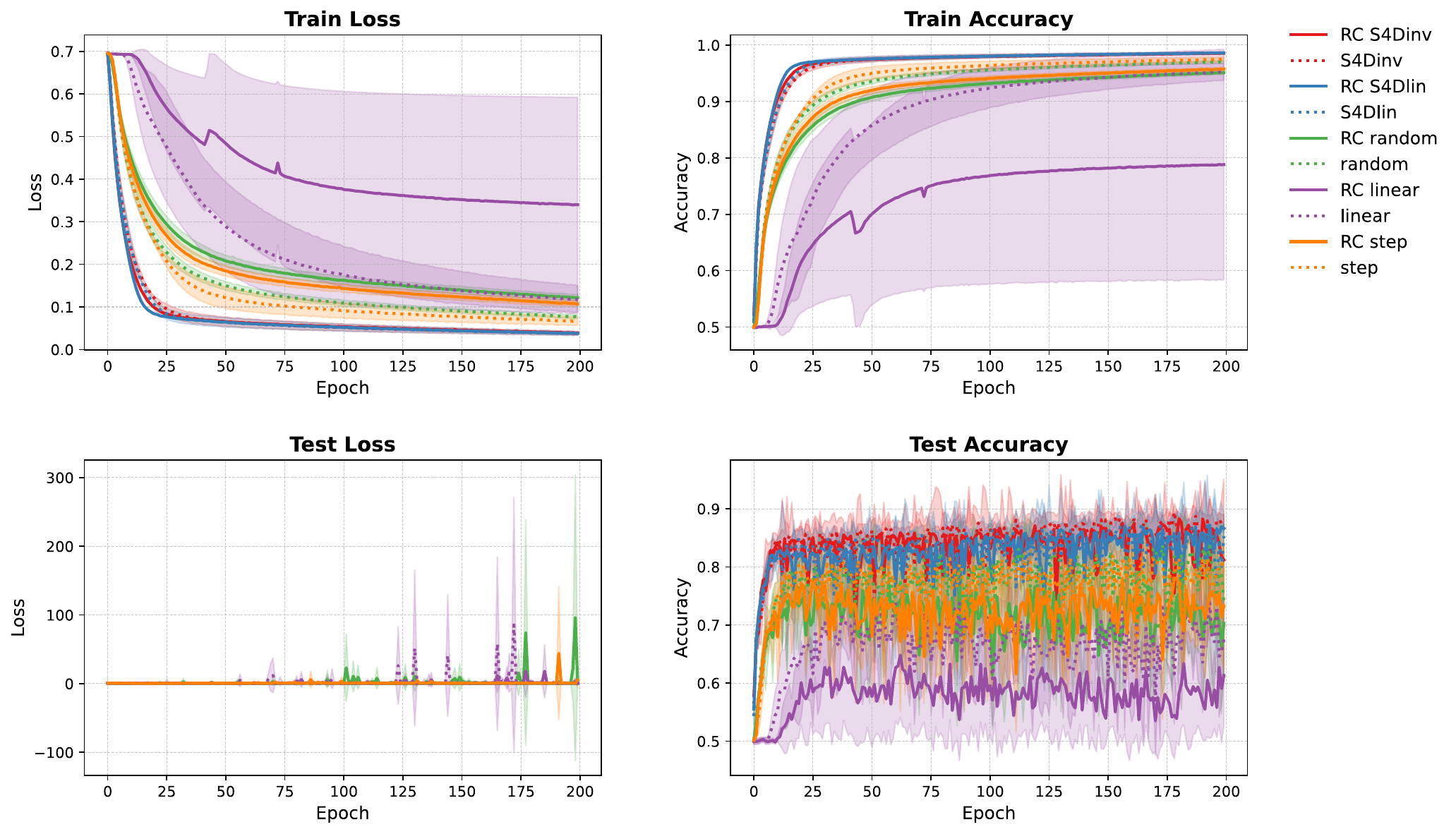}
\caption{\label{fig:acc_pathfinder}
Test loss and accuracy of during the learning of Pathfinder task.
The upper (lower) row indicates the metrics of train (test).
The column corresponds to each metric.
The horizontal axes are epochs.
The colors indicate eigenvalue realizations.
The solid lines are the model under RC setting
while the dotted lines are the model whose eigenvalues are allowed to learn.
}
\end{figure}

\begin{figure}[h]
\centering
\includegraphics[width=1.0\linewidth]{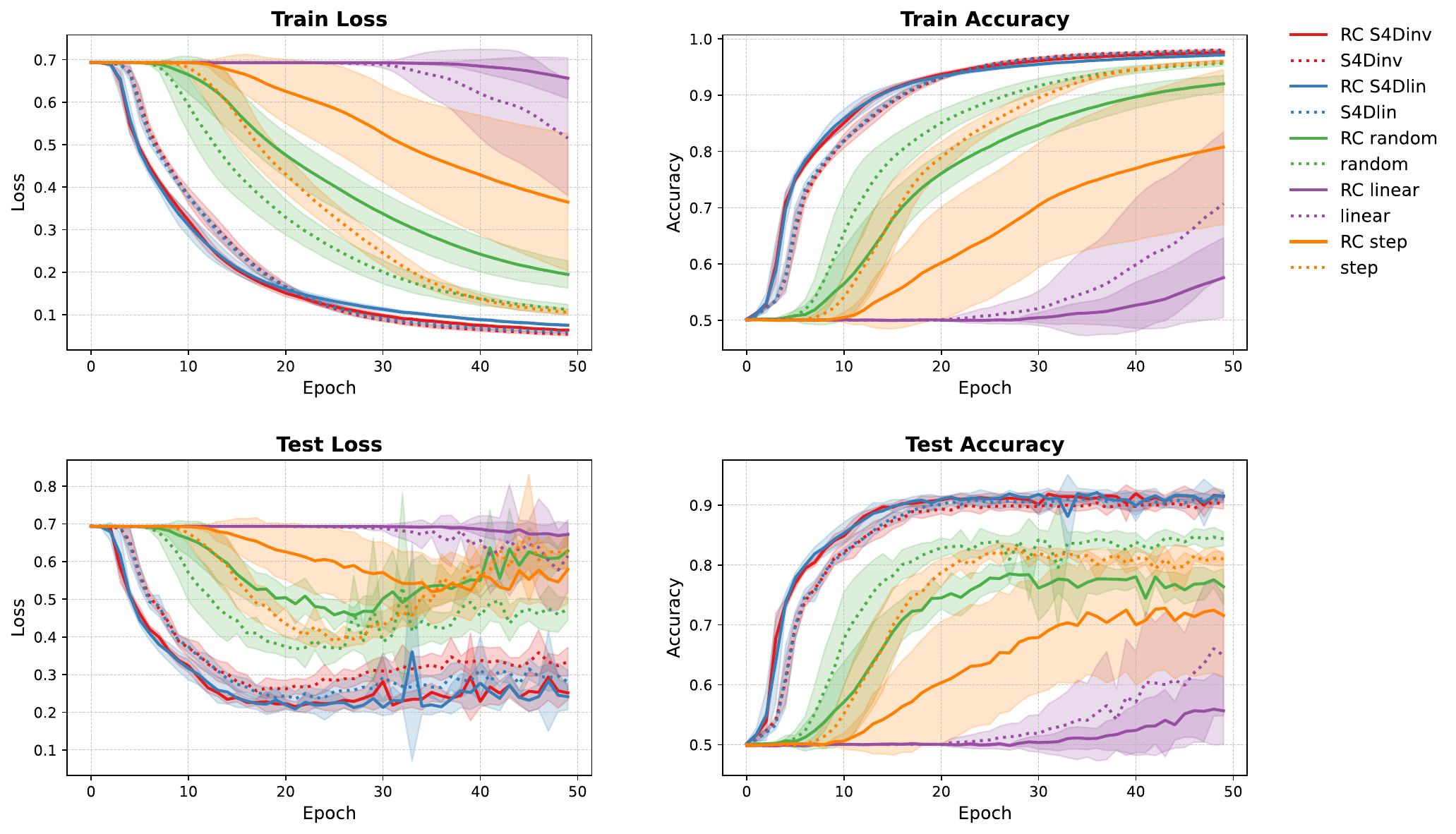}
\caption{\label{fig:acc_pathx}
Test loss and accuracy of during the learning of PathX task.
The upper (lower) row indicates the metrics of train (test).
The column corresponds to each metric.
The horizontal axes are epochs.
The colors indicate eigenvalue realizations.
The solid lines are the model under RC setting
while the dotted lines are the model whose eigenvalues are allowed to learn.
}
\end{figure}

\clearpage
\section{Further Discussion on the Applicability of Results to Broader Models}~\label{appendix:discussion}
Our study based on MF has the potential to be applied to broader models.
In particular, since the attention mechanism in Transformers can be regarded as a linear RNN,
the results derived in this work immediately enable a direct comparison between Transformers and SSMs.
An SSM possesses $N$ nodes with distinct eigenvalues, so the theoretical upper bound of its MC is $N$.
In contrast, the attention mechanism corresponds to a linear RNN with $N=1$, where the single eigenvalue is equal to $1$.
Therefore, in order for a Transformer to achieve an MC comparable to that of an SSM, it would require $N$ independent attentions.
However, if all the eigenvalues of these attention are identical,
the Vandermonde matrix which is equivalent to the states time series suffers a rank deficiency and its rank collapses to $1$.
Since the rank of the Vandermonde matrix determines the upper bound of MC in the analytical solution,
increasing the number of attentions does not improve MC.
This could be one of the reasons why SSMs can surpass Transformers.

Moreover, the indicator MF can potentially be applied to other SSMs such as Mamba and Mamba2~\citep{gu2023mamba,dao2024transformers}.
These models incorporate nonlinear operations into the RNN state updates and can therefore be regarded as nonlinear RNNs.  
For this reason, the results established in this study cannot be directly extended to them.  
Frameworks like IPC, which can accommodate nonlinearity, may offer a path forward.
In S4 and Mamba, the number of nodes $N$ takes a relatively large value (e.g., $N=32$),  
whereas in Mamba2, which is based on the Transformer architecture, we have $N=1$.  
Since both the sum of IPC and MC have the same upper bound, the state size $N$,
the significant drop in the theoretical capacity of Mamba2 is noteworthy.
This observation indicates that the high performance of Mamba2 is likely supported by the nonlinearity or other architectural elements.

\end{document}